\documentclass[Afour,sageh,times]{sagej}

\usepackage{amsmath}
\usepackage{centernot} % negation of symbols
\usepackage{amsfonts}
\usepackage{color}
\usepackage[font=small,labelfont=bf]{caption} % stylizing captions in figures
\usepackage{multirow,multicol,tabularx}
\usepackage{url}
\usepackage{verbatim} % for comments

\usepackage{mathtools}
\DeclarePairedDelimiter\abs{\lvert}{\rvert}%
\DeclarePairedDelimiter\norm{\lVert}{\rVert}%
\makeatletter
\let\oldabs\abs
\def\abs{\@ifstar{\oldabs}{\oldabs*}}
\let\oldnorm\norm
\def\norm{\@ifstar{\oldnorm}{\oldnorm*}}
\makeatother

\usepackage{bm} % bold math symbols
\newcommand{\prob}{\mathbb{P}}
\DeclareMathOperator*{\argmax}{argmax}

\DeclareMathOperator*{\expect}{\mathbb{E}}
\newcommand{\inv}{\mathcal{I}nv}

\newcommand{\entropy}{\mathrm H}

\usepackage [english]{babel}
\usepackage [autostyle, english = american]{csquotes}
\MakeOuterQuote{"}

\usepackage{graphicx}
\usepackage{epstopdf}

\usepackage[linesnumbered,ruled,vlined]{algorithm2e}
\makeatletter
\newcommand{\nosemic}{\renewcommand{\@endalgocfline}{\relax}}% Drop semi-colon ;
\newcommand{\dosemic}{\renewcommand{\@endalgocfline}{\algocf@endline}}% Reinstate semi-colon ;
\let\oldnl\nl% Store \nl in \oldnl
\newcommand{\nonl}{\renewcommand{\nl}{\let\nl\oldnl}}% Remove line number for one line
\makeatother

\usepackage{setspace}

\usepackage{tikz}
\usetikzlibrary{shapes,arrows}
\usetikzlibrary{chains}

\usepackage{amsthm}
\newtheorem{theorem}{Theorem}
\newtheorem{corollary}{Corollary}

\newtheorem{lemma}{Lemma}
\theoremstyle{definition}
\newtheorem{definition}{Definition}
\theoremstyle{remark}
\newtheorem*{remark}{Remark}

\usepackage{floatrow}
\usepackage{wrapfig}

\usepackage{subcaption}
\usepackage{empheq}

\usepackage[export]{adjustbox}

\usepackage{dblfloatfix}

\usepackage{authblk}

\usepackage{array}
\newcolumntype{?}{!{\vrule width 1.5pt}}
\usepackage{makecell}
\usepackage{empheq}
\usepackage[page,titletoc]{appendix}
\setcitestyle{authoryear,open={(},close={)}}
\graphicspath{ {./images/} }
\newcommand{\problem}{\mathcal{P}}
\newcommand{\loss}{\textit{loss}}

\begin{document}

\title{Simplified decision making in the belief~space using belief sparsification}

\author{Khen Elimelech\affilnum{1} and Vadim Indelman\affilnum{2}}
\runninghead{Elimelech and Indelman}
\affiliation{\affilnum{1}Robotics and Autonomous Systems Program, Technion –- Israel Institute of Technology.\\
             \affilnum{2}Department of Aerospace Engineering, Technion –- Israel Institute of Technology}
\corrauth{Khen Elimelech,\\ Technion, Haifa 3200003, Israel.}
\email{\texttt{khen@technion.ac.il}}

\begin{abstract}
In this work, we introduce a new and efficient solution approach for the problem of decision making under uncertainty, which can be formulated as decision making in a belief space, over a possibly high-dimensional state space. Typically, to solve a decision problem, one should identify the optimal action from a set of candidates, according to some objective. We claim that one can often generate and solve an analogous yet simplified decision problem, which can be solved more efficiently. A wise simplification method can lead to the same action selection, or one for which the maximal loss in optimality can be guaranteed. Furthermore, such simplification is separated from the state inference and does not compromise its accuracy, as the selected action would finally be applied on the original state. First, we present the concept for general decision problems and provide a theoretical framework for a coherent formulation of the approach. We then practically apply these ideas to decision problems in the belief space, which can be simplified by considering a sparse approximation of their initial belief. The scalable belief sparsification algorithm we provide is able to yield solutions which are guaranteed to be consistent with the original problem. We demonstrate the benefits of the approach in the solution of a realistic active-SLAM problem and manage to significantly reduce computation time, with no loss in the quality of solution. This work is both fundamental and practical, and holds numerous possible extensions.
\end{abstract}

\keywords{Decision making under uncertainty, belief space planning, POMDP, sparse systems, sparsification, active SLAM}

\maketitle

%\footnotetext[0]{Submitted for peer review in December 2018.}
%\tableofcontents 

\section{Introduction}
\subsection{Background}
In this era, intelligent autonomous agents and robots can be found all around us. They are designed for various functions, such as operating in remote domains, e.g., underwater and space; imitating humans and interacting with them; performing repetitive tasks; and ensuring safety of operations. They might be physically noticeable, e.g., personal-use drones, industrial robotic arms, and military vehicles; or less so, with the popularization of internet of things (IoT), smart homes, and virtual assistants. Still, these agents share the same fundamental goal -- to autonomously plan and execute their actions. Yet, the increasing demand for these "smart" systems presents new challenges: integration of robotic agents into everyday life requires them to operate in real time, using inexpensive hardware. In addition, when planning their actions, these agents should account for real-world uncertainty in order to achieve reliable and robust performance. There are multiple possible sources for such uncertainty, including dynamic environments, in which unpredictable events might occur; noisy or limited observations, such as an imprecise GPS signal; and inaccurate delivery of actions.

Also, problems, such as long-term autonomous navigation, and sensor placement over large areas, often involve optimization of numerous variables. These settings require reasoning over high-dimensional probabilistic states, known as "beliefs". Appropriately, the corresponding planning problem is known as Belief Space Planning (BSP). The objective in such a problem is to select "safe" actions, which account for the uncertainty of the agent's belief.  Other relevant instantiations include active Simultaneous Localization and Mapping (SLAM), active sensing, robotic manipulation, and even cognitive tasks, such as dialogue management.
The BSP problem is often modeled as a Partially Observable Markov Decision Process (POMDP), according to which we shall propagate the belief, and evaluate the development of uncertainty, considering \emph{multiple} courses of action \citep{Kaelbling98ai}. Further, proper uncertainty measures, such as differential entropy, are expensive to calculate for high-dimensional and continuous beliefs. Overall, the computational complexity of the problem can turn exceptionally high, thus making it challenging for online systems, or when having a limited processing power.

% In fact, the optimal solution of a POMDP was proven to be intractable \citep{Papadimitriou87math}.

\subsection{Objectives and approach overview \label{sec:intro-contributions}}
The previous discussion leads us to our main goal -- allowing computationally efficient decision making. Note that in this study, we differentiate between planning and decision making. Planning is a broad concept, which takes into consideration many aspects, such as goal setting and balancing, generation of candidate actions, accounting for different planning horizons and future developments, coordination of agents, and so on. After refining these aspects, we eventually result in a decision problem: considering an initial state, and a \emph{given} set of candidate actions (or action sequences), we use an objective function to measure the scalar values attained by applying each action on the initial state; to solve the problem, we shall identify the optimal candidate action, which generates the highest objective value. With this rudimentary view-point, we dismiss problem-specific attributes, which allows our formulation to address a wider range of problems. Nonetheless, our work heavily focuses on contributing to decision making in the belief space. In these decision problems, the initial state is a belief over a (possibly) high-dimensional state, and the objective function is a belief-based information-theoretic value, measured from the propagated (updated) belief, after applying a candidate action.

A traditional solution to the decision problem requires calculation of the objective function for each candidate action. We would like to reduce the cost of the solution by sparing this exhaustive calculation and comparison. Instead, we suggest to identify and solve a simplified decision problem, which leads to the same action selection, or one for which the loss in quality of solution can be bounded. A problem may be simplified by adapting each of its components -- initial state, objective function, and candidate actions. To allow such analysis, we first provide a general theoretical framework, which does not depend on any problem-specific attributes; the framework allows us to formally quantify the effect of the simplification on the action selection, and form optimality guarantees for it.

We then show how these ideas can be practically applied to high-dimensional BSP problems. In this case, the problem is simplified by considering a sparse approximation of the initial belief, which can be efficiently propagated, in order to calculate the candidates' objective values. The resulting simplified problem can be solved in any desired manner, making our approach complementary to other solvers. Furthermore, while several works already utilize belief sparsification to allow long-term operation and tractable state inference, the novelty in our approach is the exploitation of sparsification exclusively and dedicatedly for efficient decision making. After solving the decision problem, the selected action is then applied on the \emph{original} belief; by such, we do not compromise the accuracy of the estimated state.

For clarity, we list down the contributions of this work, in the order they are presented in the manuscript:

\begin{enumerate}
\item A theoretical framework supporting the concept of decision problem simplification;
\item Formulation of decision making in the belief space, and application of the concept to it;
\item A scalable belief sparsification algorithm;
\item Derivation of quality-of-solution guarantees;
\item Experimental demonstration in a highly realistic active-SLAM scenario, where a significant improvement in run-time is achieved.
\end{enumerate}

Please note that this paper extends our previous publications \citep{Elimelech17icra, Elimelech17isrr, Elimelech17iros}. Besides the expanded experimental evaluation, the belief sparsification algorithm, which was previously introduced, is now reformed to a more stable and efficient version. Also, the theoretical formulation includes several revisions and corrections to previously introduced definitions; the conclusive versions are those presented here. Also, to allow fluid reading, proofs for all theorems, lemmas, and corollaries are given in the appendix.

\subsection{Related work}
Several works explore similar ideas to the ones presented here. In this section we do our best to provide an extensive review of such works, in comparison to ours.

As mentioned, numerous methods consider sparsification for the probabilistic state inference problem, in order to limit the belief size, and improve its tractability for long-term operation. Although being a well-researched concept, these methods do not examine sparsification in the context of planning problems (influence over action selection, computational benefits, etc.). \cite{Thrun04ijrr}, for example, showed that in a SLAM scenario, when using the information filter, forcing a certain sparsity pattern on the belief's information matrix can lead to improved efficiency in belief update. However, they emphasized that the approximation quality was not guaranteed and that certain scenarios could lead to significant divergence.

Also, since \cite{Dellaert06ijrr} demonstrated the equivalence between sparse matrices and (factor) graphs for belief representation, graph-based solutions for SLAM problems (which is often a sparse problem) have become more popular. Accordingly, methods for graph sparsification have also gained relevance. For example, \cite{Huang13ecmr} introduced a graph sparsification method, using node marginalization. The resulting graph is notably consistent, meaning, the sparsified representation is not more confident than the original one. Several other approaches suggest to sparsify the graph using the Chow-Liu tree approximation, and show that the KL-divergence from the original graph remains low \citep{CarlevarisBianco14tro, CarlevarisBianco14icra, Kretzschmar12ijrr}. \cite{Hsiung18iros} reach similar conclusions for fixed-lag Markov blankets.
Notably, our sparsification method, which is presented both in matrix and graph forms, preserves the dimensionality of the belief, and only modifies the correlations between the variables. It is also guaranteed to exactly preserve the entropy of the belief.

The approach described by \cite{Mu17tro} separated the sparsification into two stages: problem-specific removal of nodes, and problem-agnostic removal of correlations. The authors then demonstrated the superiority of their scheme over agnostic graph optimization, in terms of collision percentage. This two-stage solution reminds the logic in our sparsification method: first, identifying variables with minimal contribution to the decision problem, and then sparsification of corresponding elements. Of course, we use such sparsification for planning and not graph optimization.

Exploiting sparsity to improve efficiency can also be done in other manners. Fundamental works \citep[e.g.,][]{Davis04toms}, alongside newer ones \citep[e.g.][]{Frey17arxiv,Agarwal12iros}, provide heuristics for variable elimination order or variable pruning order, in order to minimize fill-in during factorization of the information matrix (which is utilized during belief propagation).

In the context of planning under uncertainty and POMDP, the research community has been extensively investigating solution methods to provide better scalability for real-world problems. Finding optimal solutions (policies) according to the POMDP formulation is often done by utilizing dynamic programming algorithms, such as, value and policy iteration \citep[e.g.,][]{Porta06jmlr,Pineau06jair}. Such methods are extremely computationally demanding, especially when considering high-dimensional state space (i.e., search spaces). These methods are thus generally not suitable for "online" planning problems for autonomous agents, in which we want to infer a specific sequence of actions to be executed immediately.

Instead, when considering "online" scenarios, we typically perform a forward search from the current belief, and often forced to rely on approximated solutions. Standard online POMDP solvers \citep[e.g.,][]{Silver10nips,Ye17jair} often perform search in the state-space, and not the belief space, as we care to do here. Works which do consider planning in the belief space, typically focus on methods for alleviating the search. For example, some solution methods perform direct (localized) trajectory optimization \citep[e.g.][]{Indelman15ijrr,VanDenBerg12ijrr}. Otherwise, while building on established motions planners \citep[e.g.,][]{Karaman11ijrr, Kavraki96tra}, works such as the Belief Roadmap \citep[by][]{Prentice09ijrr}, FIRM \citep[by][]{AghaMohammadi14ijrr}, SLAP \citep[by][]{AghaMohammadi18tro}, and others \citep[e.g., by][]{Patil14wafr} rely on sub-sampling a finite graph in the belief space, in which the solution can be searched. %The graph created by these methods conveniently preserves the optimal substructure property, which allows to incrementally search for the (approximated) optimal candidate. 
However, such methods are severely limited, by only allowing propagation of the belief over a single (most-recent) pose through the graph; i.e., they perform low-dimensional pose filtering, rather than high-dimensional belief smoothing, as we do. This forced marginalization of state variables surely compromises the accuracy of the estimation, and limits the applicability to (problems such as) active-SLAM, in which we often wish to examine the information (uncertainty) of the entire posterior state, including the map and/or executed trajectory \citep{Stachniss04iros,Kim14ijrr}.

Nonetheless, we do not focus on generation (or sampling) of candidates, but, instead, on efficient comparison of their objective values, by lowering the cost of belief updates.\linebreak Hence, our approach is complementary to the aforementioned graph-based methods, which focus on generating feasible candidates. We demonstrated this compatibility in our experimental evaluation, where we used a graph-based motion planner (from the most recent pose) to simply generate a set of candidate actions; we then efficiently selected the optimal candidate by propagating the sparsified (high-dimensional) belief, and evaluating its posterior uncertainty. In that regard, we may mention additional works which similarly address the issue of high-dimensional belief propagation, in the context of active-SLAM \citep[e.g.,][]{Chaves16iros, Kopitkov17ijrr}.

%For example, it was shown \citep{Prentice09ijrr} that for a linear POMDP scenario, examining a sequence of actions can be done in a single linear calculation. Using this conclusion, trajectories to the goal, which are constructed of a sequence of sampled poses, can be examined efficiently when planning. 
%Also recent work \citep{Kopitkov17ijrr} use the matrix determinant lemma for efficient calculation of entropy-based cost function.
%In \citep{Chaves16iros} the authors demonstrate a great reduction in the calculation complexity of the information measure, by utilizing a specific variable order in the Bayes tree. This order allows sharing unchanged calculations between similar actions and successive planning iterations.

Also, closely related to our approach, several other works examine approximation of the state or the objective function in order to reduce the planning complexity. A~recent approach \citep{Bopardikar16ijrr} suggested using a bound over the maximal eigenvalue of the covariance matrix as a cost function for planning, in an autonomous navigation scenario. Benefits of using this cost function include easy computation, holding an optimal substructure property (incremental search) and the ability to account to misdetection of measurements. Yet, the actual quality of results in terms of final uncertainty, when measured in conventional methods, is unclear. Their usage of bounds in attempt to improve planning efficiency reminds aspects of our work; however, we use bounds to quantify the quality of solution. As they mention in their discussion, an unanswered question is the difference in quality of solution between planning using the exact maximal eigenvalue, and planning using its bound. Our theoretical framework might be able to provide answer to this question.

\cite{Boyen98uai} suggested maintaining an approximation of the belief for efficient state inference. This approximation is done by dividing state variables into a set number of classes, and then using a product of marginals, while treating each class of variables as a single "metavariable". A $k$-class belief simplification cuts the original exponential inference complexity by a factor of $k$. The study showed that in rapidly-mixing POMDPs the expectation of the error could be bounded. This simplification method was later examined under a restrictive planning scenario \citep{McAllester99uai}. The planning was performed using a planning-tree search, in which a constant amount of possible observations was sampled for each tree level, and again assuming a rapidly-mixing POMDP. There, the error induced by planning in the approximated belief space can be bounded as well. This method shares similar objectives with our work, but examines a very specific scenario, which limits its generality.

In the approach described by \cite{Roy05jair}, the authors attempted to find approximate POMDP solutions by utilizing belief compression, which was done with a PCA-based algorithm. This key idea is similar to ours, yet, in that work, the objective value calculation (i.e., decision making) still relied on the original decompressed belief, instead of the simplified one. Thus, no apparent computational improvement was achieved in planning complexity. The paper also did not make a comparison of this nature, and only presented analysis on the quality of compression.

The work presented by \cite{Indelman15acc, Indelman16ral} contained the first explicit attempt to use belief sparsification to specifically achieve efficient planning. The papers showed that using a diagonal covariance approximation, a similar action selection could usually be maintained, while significantly reducing the complexity of the objective calculation. This claim, however, is most often not guaranteed. Optimal action selection was only proved under severely simplifying assumptions -- when candidate actions and observations only update a single state variable, with a rank-1 update of the information. This attempt inspired our extensive research and in-depth, formal analysis.

Finally, it is worth mentioning that the idea of examining only the order of candidate actions, instead of their cardinal objective values, sometimes appears in the context of economics under the term \emph{ordinal utility} \citep[e.g.][]{Manski88tad}; this term, however, is not prominent in the context of artificial intelligence. We examine a similar idea in our theoretical framework, to follow.

\pagebreak

\section{Simplified decision making \label{sec:sdm}}
To begin with, let us consider a \emph{decision problem} $\problem$, which we formally define in Definition~\ref{dfn:dp}. 
%Note that this definition and the following discussion are general, and are independent from the previous BSP formulation.

\begin{definition}%[Decision Problem] $ $\\
A \emph{decision problem} $\problem$ is a 3-tuple $ \left({\bm\xi},\mathcal{A},V\right) $, where ${\bm\xi}$ is the \emph{initial state}, from which we examine a \emph{set of candidate actions} $\mathcal{A}$ (finite or infinite), using an \emph{objective function} $V\mathpunct{:} \{{\bm\xi}\} \times \mathcal{A} \rightarrow \mathbb{R}$. Solving the problem means selecting the \emph{optimal action} $a^*$, such that
\begin{equation}
\label{eq:decision-problem}
a^* = \argmax_{a \in \mathcal{A}} V({\bm\xi},a).
\end{equation}
\label{dfn:dp}
\end{definition}

According to our suggested solution approach, we wish to generate and solve a simplified yet analogous decision problem $\problem_s \doteq \left({\bm\xi}_s,\mathcal{A}_s,V_s\right)$, which results in the same (or similar) action selection, but for which the solution is more computationally efficient.
This can be achieved by altering or approximating any of the problem components -- initial state, candidate actions, or objective function -- in order to alleviate the calculation of the candidates' objective values. Nonetheless, approximating each of these components represents a different simplification approach. For example, there is a logical difference between simplifying the initial state (i.e., examining different states under the same objective function), and simplifying the objective function (i.e., examining the same state under different objectives); in the first case, we would like to maintain a certain relation between states, and in the second one, a relation between functions.

Next, we will introduce additional ideas to help formalize our goal, and see how these can guide us towards designing effective simplification methods, which are guaranteed to preserve the quality of solution.

\subsection{Analyzing simplifications \label{sec:analyzing-simp}}
\subsubsection{Simplification loss}
Examining a simplified decision problem may lead to loss in the quality of solution, when the selected action is not the real optimal action.
We can express this loss with the following simplification quality measure:

\begin{definition}%[Simplification Loss] $ $ \\
The \emph{simplification loss} between a decision problem $\problem \doteq \left({\bm\xi},\mathcal{A},V\right)$ and its simplified version \mbox{$\problem_s \doteq \left({\bm\xi}_s,\mathcal{A}_s,V_s\right)$}, due to sub-optimal action selection, is 
\begin{multline}
\loss(\problem,\problem_s) \doteq V({\bm\xi},a^*) - V({\bm\xi},a^*_s), \hfill\\\hfill \text{where }
a^* = \operatorname*{argmax}_{a \in \mathcal{A}} V({\bm\xi},a),\, a^*_s = \operatorname*{argmax}_{a_s \in \mathcal{A}_s} V_s({\bm\xi}_s,a_s).
\end{multline}
\label{dfn:loss}
\end{definition}

\begin{figure}
\begin{subfigure}{\textwidth}
  	\includegraphics[trim= 10 3.5 10 5, clip, width=\textwidth]{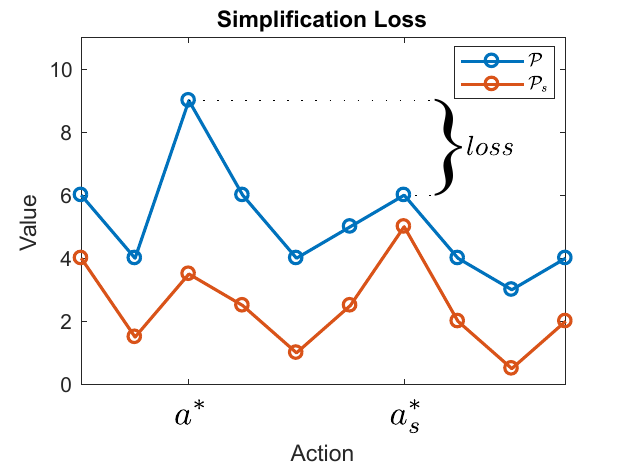} 	
  	\caption{}
\label{fig:loss-concept}
\end{subfigure}
\begin{subfigure}{\textwidth}
  	\includegraphics[trim= 10 3.5 10 3, clip, width=\textwidth]{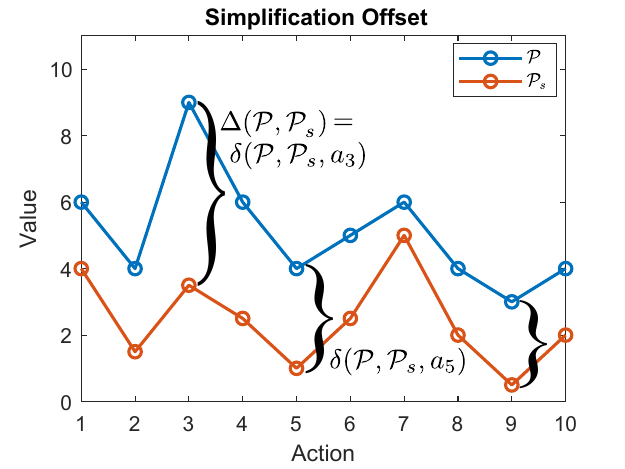} 	
  	\caption{}
\label{fig:offset-concept}
\end{subfigure}

  \caption[The simplification loss, and the simplification offset.]{$\problem_s$ is a simplified version of a decision problem $\problem$; the graphs show the objective values of each problem's candidate actions. (a) $a^*_s$ is the optimal action according to the simplified problem, and~$a^*$ is the real optimal action; the difference between the (real) objective values of these two actions is the \emph{loss} induced by the simplification.
  (b) The \emph{offset} measures the maximal difference between \emph{respective} objective values from the two problems, and does not require to explicitly identify $a^*$/$a^*_s$.}
\end{figure}

To put in words, this loss is the difference between the maximal objective value, attained by applying the optimal candidate action $a^*$ on ${\bm\xi}$, and the value attained by applying~$a^*_s$ (the action returned from the solution of the simplified solution) on ${\bm\xi}$. This idea is illustrated in Fig.~\ref{fig:loss-concept}. We implicitly assume that the original objective function~$V$ can accept actions from the simplified set of candidates~$\mathcal{A}_s$. 
When the solutions to the problems agree \mbox{$\loss(\problem,\problem_s) = 0$}.

Most often it is indeed possible to settle for simplified decision problem formulation (which can lead to a sub-optimal action), in order to reduce the complexity of action selection; though, it is important to quantify and bound the potential loss, before applying the selected action, in order to guarantee that this solution can be relied on.

\subsubsection{Simplification offset \label{sec:offset-def}}
To asses the simplification loss, we suggest to identify the \emph{simplification offset}, which acts as an intuitive "distance" measure in the space of decision problems:
\begin{definition}%[Simplification Offset] $ $ \\
\label{dfn:ro}The \emph{simplification offset} of a candidate \mbox{$a\in\mathcal{A}$}, between a decision problem $\problem \doteq \left({\bm\xi},\mathcal{A},V\right)$, and its simplified version $\problem_s \doteq \left({\bm\xi}_s,\mathcal{A},V_s\right)$ is
\begin{equation}
\delta(\problem,\problem_s,a) \doteq \abs{V({\bm\xi},a) - V_s({\bm\xi}_s,a)}.
\end{equation}
Overall, the \emph{simplification offset} between $\problem$ and $\problem_s$ is
\begin{equation}
\Delta(\problem,\problem_s) \doteq \max_{a \in \mathcal{A}} \left\{ \delta(\problem,\problem_s,a) \right\}.
\end{equation}
\end{definition}
Unlike the loss, the \emph{offset} (which is illustrated in Fig.~\ref{fig:offset-concept}) measures the maximal difference between \emph{respective} objective values from the two problems, and does not require to explicitly identify the optimal actions.
Further, for each candidate $a\in\mathcal{A}$, the offset represents an interval for the real value~$V({\bm\xi},a)$, around the respective approximated value~$V_s({\bm\xi}_s,a)$, in which it must lie, i.e.:
\begin{equation}
\label{eq:interval-sym}
V_s({\bm\xi}_s,a)- \delta(a)
 \,\leq\, V({\bm\xi},a) \,\leq\,
V_s({\bm\xi}_s,a)+ \delta(a)
\end{equation}
Notably, the offset represents only the \emph{size} of this interval, and not its \emph{location} on the value axis (around $V_s({\bm\xi}_s,a)$). %; this is demonstrated in Fig.~\ref{fig:offset-sym}.
This means that the offset, in contrast to the loss, is a property of the simplification \emph{method}, and does not depend on the solution of $\problem$~nor~$\problem_s$. It can thus potentially be examined without explicitly solving either of the problems, nor calculating $V$~nor~$V_s$, as we shall see.

Note that when defining the offset, we implicitly considered that the two problems examine the same set of candidate actions; this will be valid from now on, unless stated otherwise. Also, for brevity, we will no longer write the initial state as input to $V$/$V_s$, nor $V,V_s$ as input to $\delta/\Delta$, whenever the context is clear.
Next, we will explain how we can utilize the offset to infer loss guarantees.

\subsection{Optimality guarantees \label{sec:guarantees-general}}

\subsubsection{Bounding the offset}
Obviously, knowing the offset exactly for every action would be equivalent to having access to the original solution. We would thus usually rely on a bound of the offset to infer loss guarantees.
As mentioned, the offset measures the difference between respective objective values from the original and simplified problems, and is independent of their solutions. Thus, 
we can evaluate and attempt to bound the offset \emph{before} solving the problem;
by utilizing the general structure of problems in our domain, and knowing how they are affected by the nominative simplification method, we can try to infer a \emph{symbolic formula} for the offset, and draw conclusions from it. 
This type of analysis often allows us to draw general conclusions regarding the simplification \emph{method}, rather than a specific problem. 
For example, in Section~\ref{sec:belief-approx}, we discuss a novel belief simplification method, used to reduce the cost of planning in the "belief space". By symbolically analyzing the offset (for any decision problem in this domain), we could identify the conditions under which its value is zero, and the simplification is guaranteed to induce no loss. This idea is later demonstrated in Section~\ref {sec:dmuu-var-selection}.
Still, we note that providing completely general guarantees, which are valid for all the decision problems in the domain, is not always possible from pure symbolic analysis. Sometimes, to draw decisive conclusions, we must assign the properties of the specific decision problem we wish to solve.

If we failed to reach valuable conclusions from such "pre-solution" symbolic analysis of the offset, we can try to bound it "post-solution", by utilizing the calculated (simplified) values, and (any) known bounds, or limits, for the real objective values; these limits should be selected based on domain knowledge of the specific problem. Then, the following can be easily derived from the definition of the simplification loss:
\begin{align}
\delta(a) \leq \max\big\{ 
&V_s({\bm\xi}_s,a) - \mathcal{LB}\left\{ V({\bm\xi},a) \right\} , \nonumber\\ 
&\mathcal{UB}\left\{ V({\bm\xi},a) \right\} - V_s({\bm\xi}_s,a) \big\}
\label{eq:offset-bound-post}
\end{align}
where $\mathcal{LB},\mathcal{UB}$ stand for lower and upper bounds, respectively. 
We demonstrate how to practically utilize this idea in Section~\ref{sec:dmuu-guarantees}.

\subsubsection{Bounding the loss}
As discussed, our goal is to guarantee that relying on a certain simplification would not induce more than the acceptable loss. As with the offset, bounding the loss can be done on two occasions:
(i)~\textbf{pre-solution analysis} -- this type of analysis occurs \emph{before} solving the simplified problem (based on the availability of "symbolic" offset bounds); and (ii)~\textbf{post-solution analysis} -- which occurs \emph{after} solving the simplified problem (but before applying the selected action).
Surely, we prefer to know if the simplified solution would be worthwhile \emph{before} investing in it; for example, we may consider the case where action execution is costly (as measured with the objective function), and beyond a certain loss, improving the decision making efficiency is not worth the execution of a sub-optimal action. Nonetheless, post-solution guarantees are typically tighter, as we can also rely on the calculated values.
The notion of offset allow us to seamlessly derive both types of guarantees, and easily improve them when refining the solution, or given access to new information.

From the properties of the absolute value, it is also easy to infer that the offset is a valid metric (a distance measure) between decision problems. Indeed, Lemma~\ref{trm:loss-bound-pre} intuitively indicates that when the offset between a problem and its simplification is small, then the induced loss is also small, and the action selection stays "similar".

\begin{lemma}
\label{trm:loss-bound-pre}
For any two decision problems $\problem$ and $\problem_s$,
\begin{equation}
 0 \leq \textit{loss}(\problem,\problem_s) \leq 2\cdot\Delta(\problem,\problem_s).
\end{equation}
\end{lemma}

This conclusion is potentially reachable in pre-solution analysis, as it does not rely on the simplified solution, i.e., the calculated objective values; when these become available, in post-solution analysis, this bound can be refined, as indicated in Lemma~\ref{trm:loss-bound-post}.

\begin{lemma}
\label{trm:loss-bound-post}
For any two decision problems $\problem$ and $\problem_s$,
\begin{multline}
\label{eq:loss-bound-post}
\loss(\problem,\problem_s)  \leq \hfill\\\hfill \max\Big\{ 0 ,\, 
2\cdot\Delta(\problem,\problem_s) + \max_{a \neq a^*_s} \left\{ V_s(a)\big) \right\} - V_s(a^*_s) \Big\}.
\end{multline}
\end{lemma}

For an extended discussion regarding derivation of loss guarantees, including a proof of Lemma~\ref{trm:loss-bound-post}, and more intricate loss bounding techniques, please refer to \cite{Elimelech21thesis}. Specifically, when we do not have access to a symbolic formula for the offset, and instead rely on the "post-solution offset bound" (\ref{eq:offset-bound-post}), the expression in (\ref{eq:loss-bound-post}) simplifies to:
\begin{equation}
\label{eq:loss-bound-post-post}
\loss(\problem,\problem_s) \leq \max_{a \neq a^*_s} \left\{\mathcal{UB}\left\{ V(a) \right\}\right\} - \mathcal{LB}\left\{ V(a^*_s) \right\}.
\end{equation}
Notably, such post-solution analysis allows us to understand not only what is the maximal possible loss, but also which candidates are likely to cause it.

\subsection{Reducing simplification bias \label{sec:unbiased-simp}}
Previously, we suggested the simplification offset as a "distance measure" between decision problems, and recognized that it(s bound) can be used to bound the simplification loss. However, this distance measure may be deceiving, as the problems may appear to be separated by a large offset, even when the simplification induces a small loss. Specifically, this can be the case when the simplification causes a large "bias" in the simplified objective values.
In the following section we introduce another concept, to help us handle such scenarios.

\subsubsection{Action consistency}
\begin{figure}
\begin{subfigure}[t]{\textwidth}
  	\includegraphics[trim= 10 3.5 10 5, clip, width=\textwidth]{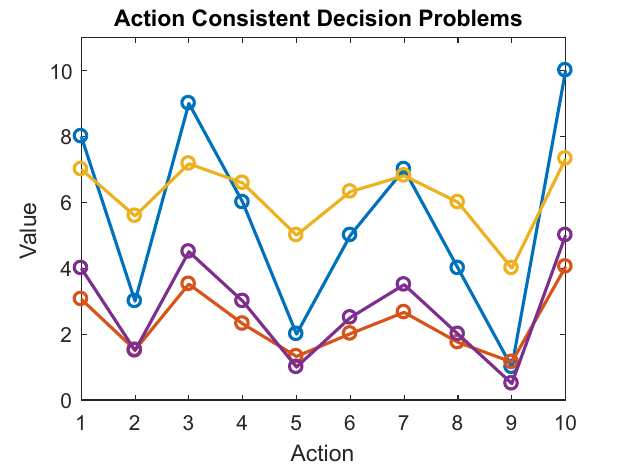} 
  	\caption{}\label{fig:ac-concept}
  \end{subfigure}
	\begin{subfigure}[t]{\textwidth}
  	\includegraphics[trim= 10 3.5 10 3.8, clip, width=\textwidth]{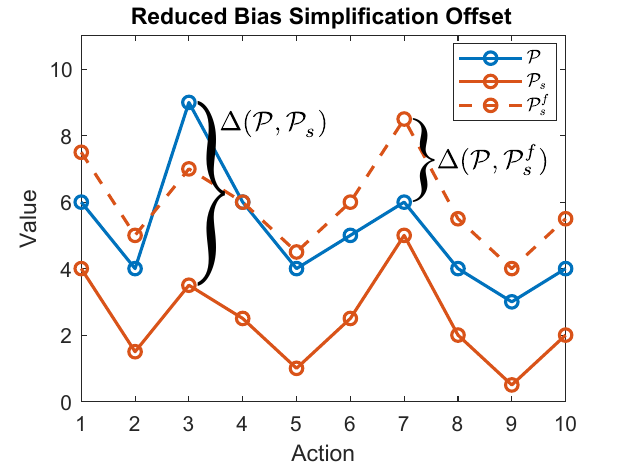} 
  	\caption{}\label{fig:unbiased-offset-concept}
  \end{subfigure}
  \caption[Action consistency, and the unbiased simplification offset.]
  {
(a) Each graph represents the objective values of the candidate actions of a certain decision problem; although the values are different, all the graphs maintain the same trend among the actions, and therefore the problems are action consistent.
(b) The simplification offset $\Delta$ between $\problem$ and $\problem_s$ is the maximal difference between the values of \emph{respective} actions. The offset can be reduced by utilizing a monotonically increasing function $f$ (here we used a constant-shift), which leads to an less biased yet action consistent problem $\problem^f_s$.
}
\end{figure}

We point out a key observation: to solve the decision problem, we only need to sort (or rank) the candidate actions in terms of their objective function value; changing the values themselves, without changing the order of actions, does not change the action selection. Hence, when two problems maintain the same order of candidate actions, their solution is equivalent. In this case, we can simply say that the two problems are \emph{action consistent}, as demonstrated in Fig.~\ref{fig:ac-concept}. 
\begin{definition}%[Action Consistency] $ $ \\
\label{dfn:ac}
Two decision problems, $\problem_1 \doteq \left({\bm\xi}_1,\mathcal{A},V_1\right)$ and $\problem_2 \doteq \left({\bm\xi}_2,\mathcal{A},V_2\right)$, are \emph{action consistent}, and marked \mbox{$\problem_1 \simeq \problem_2$}, if the following applies $\forall a_i,a_j \in \mathcal{A}$:
\begin{equation}
V_1({\bm\xi}_1,a_i) < V_1({\bm\xi}_1,a_j) \iff V_2({\bm\xi}_2,a_i) < V_2({\bm\xi}_2,a_j).
\end{equation}
If also $V_1 \equiv V_2$, we can simply say that ${\bm\xi}_1,\, {\bm\xi}_2$ are \emph{action consistent}, and mark \mbox{${\bm\xi}_1 \simeq {\bm\xi}_2$}.
\end{definition}

This relation holds several interesting properties.

\begin{lemma}% $ $ \\
\label{trm:ac-equiv-relation}
Action consistency ($\simeq$) is an equivalence relation; i.e., any three decision problems $\problem_1, \problem_2, \problem_3$, satisfy the following properties:%\setlist{nolistsep}
\begin{enumerate}%[noitemsep]
\item Reflexivity: $ \,\problem_1\simeq\problem_1$.
\item Symmetry: $ \,\problem_1\simeq\problem_2 \iff \problem_2\simeq\problem_1 $.
\item Transitivity: $ \,\problem_1\simeq\problem_2 \wedge \problem_2\simeq\problem_3 \,\Longrightarrow\, \problem_1\simeq\problem_3$.
\end{enumerate}
\end{lemma}

Lemma~\ref{trm:ac-equiv-relation} implies that the entire space of decision problems is divided into separate equivalence-classes of action consistent problems. Lemma~\ref{trm:ac-iff-f} adds that we can transfer between action consistent problems using monotonically increasing functions. We remind again that all proofs are given in Appendix~\ref{apndx:proofs}.

\begin{lemma}% $ $\\
\label{trm:ac-iff-f}
For any two decision problems $\problem_1$ and $\problem_2$,
\begin{multline}
\problem_1 \simeq \problem_2 \, \iff \text{ the mapping } f\mathpunct{:}\, V_1({\bm\xi}_1,a) \mapsto V_2({\bm\xi}_2,a) \\\hfill\text{ is monotonically increasing.}
\end{multline}
\end{lemma}

Meaning, if the (scalar) mapping of respective objective values between the two problems agrees with a monotonically increasing function (e.g., a constant shift, a linear transform, or a logarithmic function), then the problems are action consistent. If this mapping is not monotonically increasing, then the problems are not action consistent.

\subsubsection{Unbiased simplification offset}
The notion of action consistency can help us to achieve better guarantees when utilizing our previously developed analysis approach. We now understand that when deriving loss bounds, instead of examining a simplified problem $\problem_s$, we can, equivalently, examine any other problem $\problem^f_s$ that is action consistent with it. Further, such a problem will necessarily be of the form $\problem^f_s \doteq \left({\bm\xi}_s,\mathcal{A},f \circ V_s\right)$, where $f$ is monotonically increasing.

Accordingly, instead of examining the simplification offset, as considered thus far, we can examine the \emph{unbiased simplification offset}:

\begin{definition}%[Simplification Offset] $ $ \\
\label{dfn:ro-unbiased}
The \emph{unbiased simplification offset} between a decision problem $\problem \doteq \left({\bm\xi},\mathcal{A},V\right)$, and its simplified version $\problem_s \doteq \left({\bm\xi}_s,\mathcal{A},V_s\right)$ is
\begin{multline}
\Delta^*(\problem,\problem_s) \doteq \min \left\{ \Delta(\problem,\problem^f_s) \,\mid\, f\mathpunct{:}\, \mathbb{R} \rightarrow \mathbb{R} \right.\\\hfill\left. \text{ is monotonically increasing } \wedge\, \problem^f_s \doteq \left({\bm\xi}_s,\mathcal{A},f \circ V_s\right)\right\}.
\end{multline}
\end{definition}
The unbiased offset is the minimal offset between $\problem$ and any problem action consistent with~$\problem_s$. A demonstrative example appears in Fig.~\ref{fig:unbiased-offset-concept}.
Specifically, $\problem \simeq \problem_s$, if and only if the unbiased offset is zero:
\begin{lemma}
\label{trm:ro-iff-ac}
For any two decision problems $\problem$ and $\problem_s$,
\begin{equation}
\problem \simeq \problem_s \iff \Delta^*(\problem,\problem_s) = 0.
\end{equation}
\end{lemma}

Thankfully, our previous conclusions still hold, and  we can use the unbiased simplification offset to bound the loss:
\begin{lemma}
\label{trm:loss-bound-unbiased}
For any two decision problems $\problem$ and $\problem_s$,
\begin{equation}
 0 \leq \loss(\problem,\problem_s) \leq 2\cdot\Delta^*(\problem,\problem_s).
\end{equation}
\end{lemma}
Since \mbox{$ \Delta^*(\problem,\problem_s) \leq \Delta(\problem,\problem^f_s) $}, for \emph{any} monotonically increasing~$f$. We can symbolically develop $\Delta(\problem,\problem^f_s)$, for \emph{any} such~$f$ that is convenient, in order to bound the loss; such a function should help "counter" the effect of the simplification on the objective values. 
We may also recognize that the unbiased offset satisfies the triangle inequality (like the standard offset):
\begin{lemma}% $ $\\
\label{trm:ro-triang}
For any three decision problems $\problem_1$, $\problem_2$, and~$\problem_3$, the unbiased simplification offset satisfies the triangle inequality, i.e.,
\begin{equation}
\Delta^*(\problem_1,\problem_2) + \Delta^*(\problem_2,\problem_3) \geq \Delta^*(\problem_1,\problem_3).
\end{equation}
\end{lemma}

This property can potentially help in bounding the loss, when applying multiple simplifications.
However, unlike the standard offset, the unbiased offset is scaled according to the original objective values (like the loss), and is asymmetric in its input arguments. It is, therefore, not considered a metric\footnote{Still, the aforementioned properties, along with the obvious non-negativity, make the unbiased offset a quasi-metric (or asymmetric metric), which induces an appropriate topology on the space of decision problems, as explained by \cite{Kunzi01book}.}.

%\paragraph*{Related concepts}
We may also note that the notions of action consistency and simplification offset are related to the concept of "rank correlation" -- a scalar statistic which measures the correlation between two ranking vectors \citep[see][]{Kendall48book}. Yet, such ordinal vectors are oblivious to the cardinal objective values, and, therefore, cannot be used to bound the simplification loss. The rank correlation coefficient mostly serves for statistical analysis, as its calculation requires perfect knowledge on the ranking vectors. Since the rank variables are not independent of each other, a change or addition of a single vector entry may subsequently lead to change in all other entries, and require complete recalculation of the correlation coefficient. On the other hand, the concepts we introduced rely on a "local relation" between the problems: to check for action consistency, we only examine pairs of actions at a time; and to evaluate the offset -- only pairs of respective objective values. Addition of candidates, for example, does not affect these relations between the existing candidates. As we explain next, this locality can be utilized to derive offset and loss bounds.

%\section{Simplified Decision Making \label{sec:sdm}}
%\input{approach}

\section{Decision making in the belief space \label{sec:bsp}}
In the previous section, we examined the concept of decision problem simplification. We now wish to practically apply this idea to allow efficient decision making under uncertainty, which we formulate as decision making in the belief space. In this domain, the initial state of the decision problem is actually a probability distribution ("belief"), and, as to be explained, the problem is simplified by considering a sparse approximation of it. We provide an appropriate sparsification algorithm, and then show that the induced loss can be bounded. First of all, we define the problem.

\subsection{Problem definition \label{sec:bsp-def}}
\subsubsection{Belief propagation}
We consider a sequential probabilistic process. At time-step~$k$, an agent transitions from pose $x_{k-1}$ to pose $x_k$, using a control $u_k$. It then receives an observation of the world $z_k$, based on its updated state. The agent's state vector $\bm X_k \doteq (x^T_0,\dots,x^T_k,\bm L^T_k)^T$ consists of the series of poses, and may also include external variables, which are introduced by the observations; for example, in a full-SLAM scenario, $\bm L_k$ can stand for the positions of maintained landmarks.

Pose transition and observation are both probabilistic operations, which induce probabilistic constraints over the state variables, known as factors.
Here, we assume the transition and observation models are described with the following dependencies:
\begin{align}
\label{eq:model-trans}
x_k &= g_k(x_{k-1},u_k) + w_k,\quad &w_k \sim \mathcal{N}(0, \bm W_k),\\
\label{eq:model-obs}
z_k &= h_k(\bm X_k) + v_k,\quad &v_k \sim \mathcal{N}(0, \bm V_k),
\end{align}
where $\bm W_k, \bm V_k$ are the covariance matrices of the respective normally-distributed (Gaussian) zero-mean noise models $w_k, v_k$, and $g_k, h_k$ are deterministic functions. 

At each time-step, the agent maintains the posterior distribution over its current state vector $\bm X_k$, given the controls and observations taken until that time; this distribution, which is defined by the product of these factors, is also known as its \emph{belief}:
\begin{equation}
\label{eq:belief-factors}
b_k \doteq \prob(\bm X_k \mid u_{1:k}, z_{1:k}) \propto \prod_{i=1}^k f^{u_i} f^{z_i} ,
\end{equation}
where $u_{1:k} \doteq \{ u_1,\ldots, u_k\}$ and $z_{1:k} \doteq \{ z_1,\dots, z_k\}$, and $f^{u_i},f^{z_i}$ are the factors matching the respective controls and observations.
As widely considered, by utilizing local model linearization, we may conclude that given the previously-defined models, the belief $b_k$ is also normally-distributed (for the full derivation see \cite{Elimelech21thesis}).
Hence, to describe it, we can use a covariance matrix~$\bm \Sigma_k$, or equivalently, its inverse, the (Fisher) information matrix~$\bm \Lambda_k$:
\begin{equation}
\label{eq:belief}
b_k = \mathcal{N}\left({\bm X}^*_k, \bm \Sigma_k\right) \equiv \mathcal{N}\left({\bm X}^*_k, \bm \Lambda_k^{-1}\right).
\end{equation}
The matrices are symmetric, and the order of their rows and columns matches the specific order of variables in the state.

We may now reason about a posterior belief $b_{k+1}$, after performing a control $ u_{k+1}$ and taking an observation $ z_{k+1}$:
\begin{multline}
\label{eq:belief-update}
b_{k+1} \doteq \prob(\bm X_{k+1} \mid u_{1:k+1}, z_{1:k+1}) \propto \\
b_k \cdot \prob( x_{k+1} \mid  x_k, u_{k+1}) \cdot \prob( z_{k+1} \mid \bm X_{k+1}).
%&\propto \exp\left[-{1\over2} \left(\norm{X_k-\hat X_k}^2_{\Lambda^{-1}_k} + \norm{x_{k+1}-g(x_k,a_k)}^2_{W_k} + \norm{z_{k+1}-h(X_{k+1})}^2_{V_{k+1}} \right)\right].
\end{multline}
%where $\norm{X}_\Sigma \doteq \sqrt{X^T\Sigma^{-1} X}$ is the standrd notation for the Mahalanobis distance.
This belief remains normally-distributed and can be described with the following information matrix:
\begin{equation}
\label{eq:info-plus-jacob-full}
\bm \Lambda_{k+1} = \breve{{\bm \Lambda}}_k + \bm G_{k+1}^T \bm W_{k+1}^{-1} \bm G_{k+1} + \bm H_{k+1}^T \bm V_{k+1}^{-1} \bm H_{k+1},
\end{equation}
where the matrices $\bm G_{k+1}$ and $\bm H_{k+1}$ are the Jacobians $\nabla g_{k+1} \vert_{\overline {\bm X}_{k+1}}$ and $\nabla h_{k+1} \vert_{\overline {\bm X}_{k+1}}$, respectively, around some initial estimate, and $\breve{\bm \Lambda}_k$ is the augmented prior information matrix.
Since controls and observations may introduce new variables to the state vector, its size at time-step $k$, often does not match its size at time-step $k+1$. Hence, the prior information matrix $\bm \Lambda_k$ should be augmented to accommodate these new variables. We use the accent $\breve{\,\square\,}$ to indicate augmentation of the prior information matrix (with entries of zero) to match the posterior size. Adding new variables is possible at any index in the state, as long as we make sure the augmentation keeps the same variable order. If the prior state is of size $n$, and we add $m$ new variables to the end of it, then
\begin{equation}
\label{eq:prior-aug}
\breve{{\bm \Lambda}}_k \doteq 
\left(\begin{array}{c|c}
\bm \Lambda_k^{n \times n} & \bm 0^{n \times m} \\ \hline \bm 0^{m \times n} & \bm 0^{m \times m}
\end{array}\right).
\end{equation}

The expression in (\ref{eq:info-plus-jacob-full}) can be written in a more compact form, by marking the \emph{collective Jacobian} $\bm J^\delta_{k+1}$, which encapsulates the new information regarding the control and the succeeding observation:
\begin{multline}
\label{eq:info-plus-jacob}
\bm \Lambda_{k+1} = \breve{{\bm \Lambda}}_k + {\bm J^\delta_{k+1}}^T \bm J^\delta_{k+1},\hfill\\\hfill \text{where } \bm J^\delta_{k+1} = \begin{bmatrix}
    \bm W_{k+1}^{-\frac12} \bm G_{k+1} \\ \bm V_{k+1}^{-\frac12} \bm H_{k+1}
\end{bmatrix}.
\end{multline}
Each belief update can be described using a collective Jacobian of this form. Thanks to the additivity of the information, we can easily examine the information matrix of the posterior belief $b_{k+T}$ after applying a sequence of $T$ controls $u\doteq u_{k+1:k+T}$; the respective collective Jacobians of each control can simply be stacked to yield the collective Jacobian $\bm U$ of the entire sequence $u$:
\begin{multline}
\label{eq:info-additivity}
\bm \Lambda_{k+T} = \breve {\bm \Lambda}_k + \sum_{t=1}^{T} {\bm J^\delta_{k+t}}^T \bm J^\delta_{k+t} \doteq \breve{\bm \Lambda}_k + \bm U^T \bm U,
\hfill\\\hfill\text{where }
\bm U \doteq
\begin{bmatrix}
    \bm J^\delta_{k+1} \\ \vdots \\ \bm J^\delta_{k+T}
\end{bmatrix}.
\end{multline}

\subsubsection{Decision making}
At time-step~$k$, the agent performs a planning session.
According to its current (prior) belief~$b_k$, it wishes to select the control sequence which minimizes the expected uncertainty in the future (posterior) belief. To measure the uncertainty we use the differential entropy, which, for a normally-distributed belief $b$ of state size $n$, with an information matrix $\bm \Lambda$, is
\begin{equation}
\label{eq:entropy}
\entropy(b) = \frac12\cdot\ln\left[\frac{(2\pi e)^{n}}{\abs{\bm \Lambda}}\right] = -\frac12 \cdot\left( \ln\abs{\bm \Lambda} - n\cdot\ln (2\pi e) \right),
\end{equation}
where $\abs{\square}$ represents the determinant operation. 
Although other uncertainty measures with a lower computational cost exist, e.g., the trace of the covariance matrix, the entropy bests those by taking inter-variable correlations into account; those can have a dramatic effect on the measured uncertainty, and are crucial for correct analysis.
Thus, while utilizing the information update rule from (\ref{eq:info-additivity}), we define the following information-theoretic value or objective function, which measures the expected information gain between the
current and final beliefs:
\begin{equation}
\tilde{V}(b_k, u) \doteq \expect_\mathcal{Z} \left[ \entropy(b_k)-\entropy(b_{k+T}) \right],
\end{equation}
where $u$ is a candidate control sequence, and $\mathcal{Z}$ is the set of observations taken while performing this sequence.
We may also take the common assumption of achieving the most likely observations, around the current mean ("maximum likelihood" assumption, as examined by \cite{Platt10rss}), which would allow us to drop the expectation from this expression. We will also drop the augmentation mark and time index from now on, for the sake of concise writing.

Overall, from an initial belief $b$, and considering a given set of candidate control sequences $\mathcal{U}$, we are interested in solving the decision problem $\problem \doteq (b,\mathcal{U},V)$, where $V$ is the objective function%\footnote{Certainly, the determinant of the posterior information matrix $\bm \Lambda + \bm J^T \bm J$ is non-negative (positive semi-definite matrix), and larger than this of the prior (as explained in Eq.~\ref{eq:minkowski}); however, the objective function may still yield values which are negative, or lower than the "prior value", due to the normalization element.}
:
\begin{equation}
\label{eq:objective}\boxed{
V(b, u) \doteq \frac12 \cdot\left( \ln \abs{\breve{{\bm \Lambda}} + \bm U^T \bm U} - \ln \abs{{\bm \Lambda}} - m\cdot\ln (2\pi e) \right),
}\end{equation}
$\bm \Lambda$ is the information matrix of the prior belief $b$, $\bm U$ is the collective Jacobian of $u$, and $m$ is the number of variables added to the state when executing $u$ (the difference between the number of columns in $\bm U$ and in $\bm \Lambda$). 

For clarification, we described the process as sequential to conform to the common POMDP framework; we treat every planning session as a separate decision problem. 
Further, the "maximum likelihood" assumption is not essential, but is used to achieve a clear discussion, where each candidate control sequence can be described with a single collective Jacobian; for a generalized discussion, where this assumption is relaxed, and where we also allow examination of candidate policies, please see \cite{Elimelech21thesis}. Finally, we can use the information matrix to examine the future beliefs, even if the state inference process is not based on such information smoother. If the initial information matrix is not provided, it can be calculated by inverting the covariance matrix.

\subsubsection{The square root matrix \label{sec:root-form}}
An alternative way to represent the belief $b_k$ (and propagate it), is using the upper triangular square root matrix ${\bm R}_k$ of the information matrix ${\bm \Lambda}_k$, given (e.g.) by calculating the Cholesky factorization:
\begin{equation}
\label{eq:root-def}
\bm \Lambda_k = \bm R_k^T \bm R_k.
\end{equation}
Like $\bm \Lambda_k$, the order of rows and columns of ${\bm R}_k$ also matches the  order of variables in the state. Prominent \mbox{state-of-the-art} SLAM algorithms, e.g., iSAM2 \citep{Kaess12ijrr}, rely on this representation, as it allows the calculation of the posterior mean (state inference) to be performed incrementally, while exploiting inherent sparsity.

Our belief simplification method, as described in the following section, also relies on this representation. Unfortunately, in this form, the information update losses its convenient additivity property, and requires re-calculation (or update) of the factorization, in order to find the posterior square root matrix $\bm R_{k+T}$, such that
\begin{equation}
\label{eq:root-update}
{\bm R}_{k+T}^T {\bm R}_{k+T} = \bm \Lambda_{k+T} = \\ \breve{{\bm R}_k}^T \breve{{\bm R}_k} + \bm U^T \bm U, %\equiv \begin{pmatrix} \breve {\bm R}_k \\ \bm U \end{pmatrix}^T \begin{pmatrix} \breve {\bm R}_k \\ \bm U \end{pmatrix},
\end{equation}
where $\bm U$ is defined as in (\ref{eq:info-additivity}), and $\breve{{\bm R}_k}$ marks an appropriate augmentation of the prior root matrix:
\begin{equation}
\breve{{\bm R}_k} \doteq 
\left(\begin{array}{c|c}
\bm R_k^{n \times n} & \bm 0^{n \times m} \end{array}\right).
\end{equation}
On the other hand, the determinant of the posterior information can be calculated in linear time -- by multiplying of the diagonal elements of this triangular matrix.
The objective function (\ref{eq:objective}) can thus be re-written as 
\begin{multline}
\label{eq:objective-root}
V(b, u) \equiv \hfill\\\frac12 \cdot\left( \sum_{i=1}^N \ln (\bm R^+_{ii} )^2 - \sum_{i=1}^n \ln (\bm R_{ii} )^2 - m\cdot\ln (2\pi e) \right),
\end{multline}
where $n$ is the prior state size, $N$ is the posterior state size, $\bm R^+$ marks the posterior square root matrix, and the subscript $\square_{ij}$ marks the matrix element in the $i$-th row and $j$-th column.
As explained, using this form, the significant computational cost of calculating the objective value moves from the determinant calculation to the information update phase, though this can be performed incrementally.

\subsection{Belief sparsification \label{sec:belief-approx}}
We now wish to present a simplification method for the decision problem we have just formalized: $\problem \doteq \left(b,\mathcal{U},V\right)$. We choose keep the same objective function $V$, and set $\mathcal{U}$ of candidate actions, and focus on simplifying the initial belief $b$. As stated, candidate actions here are actually control sequences for the agent; we assume the collective Jacobians for the set of actions are available.

As we saw, calculation of the objective function (as defined in (\ref{eq:objective})) involves calculation of the determinant of the posterior information matrix, after performing an appropriate belief update for the candidate action. %A single determinant calculation of a matrix of size $n \times n$ is valued at $O(n^3)$ at worst.
The cost of this calculation depends directly on the number of non-zero elements in the matrix, and is significantly lower for sparse matrices. %Thus, having sparser posterior information matrices shall reduce the cost of solving the decision problem, as desired.
Thanks to the additivity of the information, sparsifying the prior information matrix $\bm \Lambda$ could potentially lead to a sparser posterior information matrix $\bm \Lambda + \bm U^T \bm U$, for every candidate action $u$ with collective Jacobian $\bm U$; notably, such sparsification of the prior is only calculated once, for any number of actions. We also note that in many problems, especially in navigation problems, the collective Jacobians are inherently sparse, and as the state grows, involve less variables in relation to its size. Hence, even after their addition to the sparsified prior information matrix, its sparsity shall be retained. Equivalently, we may seek to sparsify $\bm R$, the square root of $\bm \Lambda$, which is used in (\ref{eq:objective-root}), in order to improve the efficiency of the factorization update process.

\begin{figure}[H]\center
% Define block styles
\makeatletter
\tikzset{west above/.code=\tikz@lib@place@handle@{#1}{south west}{0}{1}{north west}{1}}
\tikzset{west below/.code=\tikz@lib@place@handle@{#1}{north west}{0}{-1}{south west}{1}}
\tikzset{east above/.code=\tikz@lib@place@handle@{#1}{south east}{0}{1}{north east}{1}}
\tikzset{east below/.code=\tikz@lib@place@handle@{#1}{north east}{0}{-1}{south east}{1}}
\makeatother
\tikzstyle{io} = [trapezium, trapezium left angle=80, trapezium right angle=100, minimum height=1.4cm, text centered, draw=black, fill=blue!40]
\tikzstyle{block} = [rectangle, draw, fill=blue!40, text centered, rounded corners, minimum height=2em]
\tikzstyle{block-opt} = [rectangle, draw, fill=blue!10, text centered, rounded corners, minimum height=2em]
\tikzstyle{line} = [draw, -latex']
\begin{tikzpicture}[start chain, node distance = 2.5mm, auto]
    % Place nodes
    \node [block-opt=->,on chain=going below, align=center] (identify) {Identify uninvolved variables};
    \node [block=->,on chain=going below, align=center] (s) {Select a subset $\mathcal{S}$ of state variables to sparsify};
    \node [block=->,on chain=going below, align=center] (find) {Find a sparse approximation $b_s$ of \\ the initial belief using Algorithm~\ref{alg:sparsification}};
    \node [block-opt=->,on chain=going below, align=center] (pre) {Pre-solution analysis};
    \node [block=->,on chain=going below, align=center] (calc) {Calculate the objective values \\ for all candidates using $b_s$};
    \node [block=->,on chain=going below] (select) {Select the "optimal" candidate};
    \node [block-opt=->,on chain=going below, align=center] (bound) {Post-solution analysis:\\ derive loss bounds, to guarantee the quality of solution};
    \node [block=->,on chain=going below] (apply) {Apply the selected action on the \emph{original} belief $b$};
    \node [io=->,above=of identify.170, text width=1.6cm] (init) {Initial belief $b$};
    \node [io=->, right=of init, text width=4cm] (jacobians) {Updates corrsponding to each candidate action ("collective Jacobians")};
    % Draw edges
    \path [line] (init) -- (identify);
    \path [line] (jacobians) -- (identify);
    \path [line] (identify) -- (s);
    \path [line] (s) -- (find);
    \path [line] (find) -- (pre);
    \path [line] (pre) -- (calc);
    \path [line] (calc) -- (select);
    \path [line] (select) -- (bound);
    \path [line] (bound) -- (apply);
\end{tikzpicture}
%vspace{-2pt}
\caption[A flowchart: belief sparsification for efficient decision making in the belief space.]{Belief sparsification for efficient decision making in the belief space. Essential steps are in dark blue; optional steps, in order to provide guarantees, are in light blue. Here, candidate actions represent control sequences for the agent.}
\label{fig:flowchart}
\end{figure}
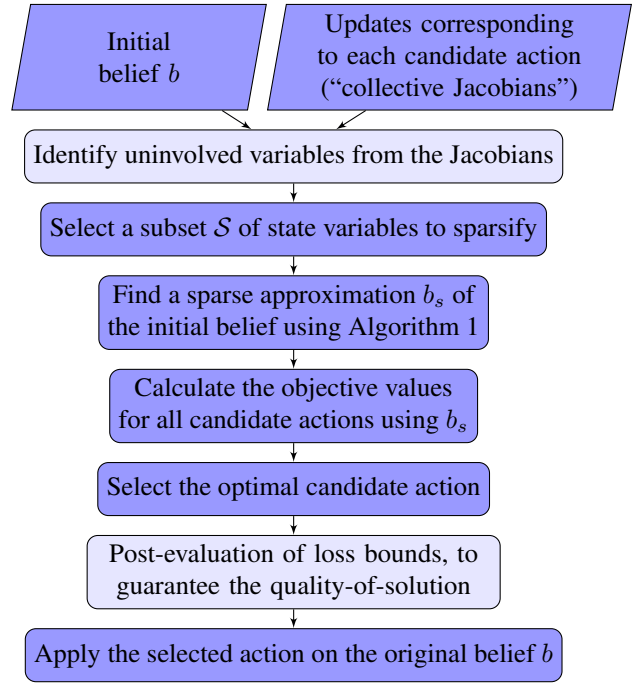

Overall, assuming the initial belief of the decision problem is $b = \mathcal{N}(\bm X^*,\bm \Lambda^{-1})$, our simplified problem shall rely instead on $b_s = \mathcal{N}(\bm X^*,\bm \Lambda_s^{-1})$ as the initial belief, where $\bm \Lambda_s$ is a sparse approximation of $\bm \Lambda$. 
In the following section, we present a sparsification algorithm\footnote{Algorithm~\ref{alg:sparsification} is a revised version of the sparsification algorithm that appeared in our previous publication \citep{Elimelech17iros}.} for the information matrix (or its square root matrix). Fig.~\ref{fig:flowchart} summarizes the paradigm of belief sparsification for efficient decision making in the belief space; clarification regarding its steps is to follow.

\subsubsection{The algorithm \label{sec:algo}}
\begin{figure*}[t]\center
  \includegraphics[width=\textwidth]{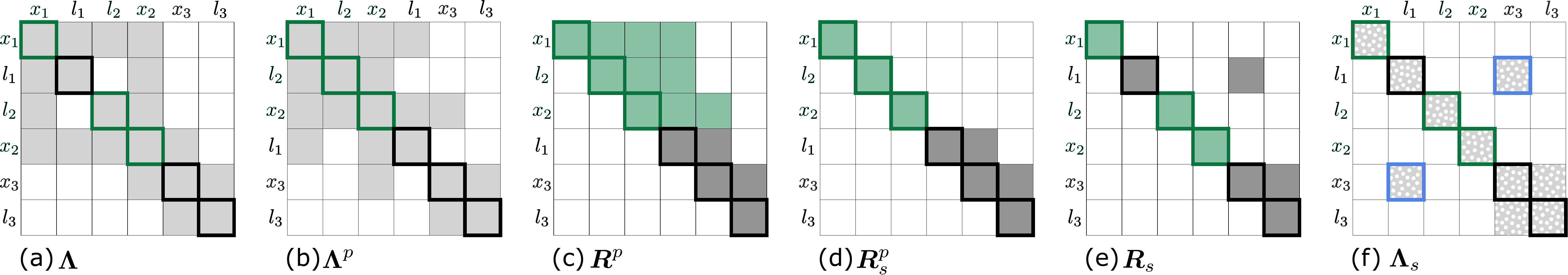}
  \caption[Belief sparsification -- matrix view.]{The steps of Algorithm~\ref{alg:sparsification} (from left-to-right), for sparsification of a Gaussian belief (shown in Fig.~\ref{fig:sparsification-graphs}a); the state variables are $\bm X \doteq [ x_1,l_1,l_2,x_2,x_3, l_3 ]^T$ (in that order), and the subset of variables selected for sparsification is $\mathcal{S} = \{ x_1,l_2,x_2 \}$ (in green).
  (a) The sparsity pattern of the symmetric information matrix of belief.
  (b) Reordering the variables, such that all the variables in $\mathcal{S}$ appear first; this is done by simply permuting the rows and columns of the matrix.
  (c) Calculating the upper triangular square root matrix $\mathtt{chol}({\bm \Lambda}^p)$ of the permuted information matrix; each row corresponds to a state variable.
  (d) Removing off-diagonal elements from rows corresponding to variables in $\mathcal{S}$.
  (e) After the sparsification, we may permute the variables back to their original order directly in the square root matrix, without breaking its upper triangular shape.
  (f) Reforming the sparsified information matrix ${\bm \Lambda}_s \doteq {\bm R}_s^T {\bm R}_s$; note that the process affects the values in the matrix, and may also introduce new non-zeros (marked in purple).}
  \label{fig:sparsification-matrices}
\end{figure*}

\begin{figure*}[t]\center
  \includegraphics[width=\textwidth]{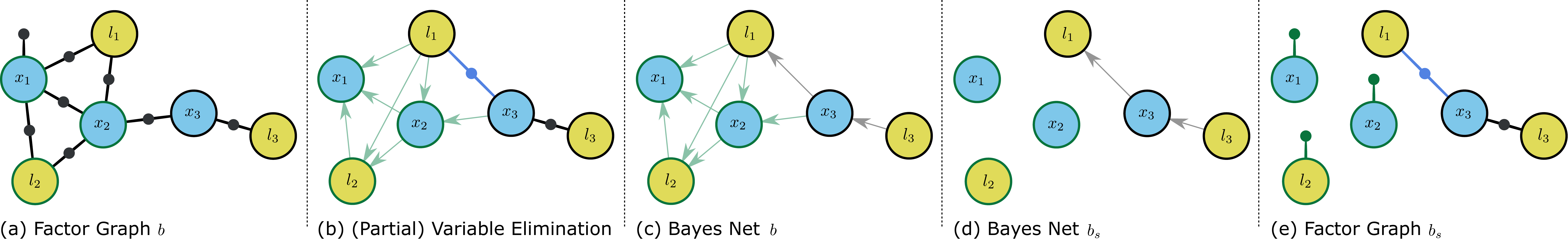}
  \caption{Visualizing the steps of Algorithm~\ref{alg:sparsification} (from left-to-right), for sparsification of a belief with probabilistic graphical models. \linebreak
  (a) The factor graph of the prior belief $b$ (matching Fig.~\ref{fig:sparsification-matrices}a); the state variables are $\bm X \doteq [ x_1,l_1,l_2,x_2,x_3, l_3 ]^T$, and the subset of variables selected for sparsification are $\mathcal{S} = \{ x_1,l_2,x_2 \}$ (circled in green).
  (b) Eliminating the variables in the factor graph in order to derive the corresponding Bayes net; the figure describes an intermediate step of the elimination process, after eliminating the variables in $\mathcal{S}$: $x_1,l_2,x_2$ (in this order); note the added marginal factor (in purple).
  (c) The final Bayes net of $b$, after eliminating all the variables.
  (d) Removing all edges which lead to variables in $\mathcal{S}$ (green arrows); this is the Bayes net describing the sparsified belief $b_s$.
  (e) Reforming the factor graph of the sparsified belief $b_s$; variables in $\mathcal{S}$ are now independent, and each is connected to a modified prior factor (in green); the remaining variables are inter-connected with the same factors which connected them originally (in black), alongside the marginal factors, which were added after elimination of $\mathcal{S}$ (in purple). }
  \label{fig:sparsification-graphs}
\end{figure*}

\begin{figure}[b]\center
  \includegraphics[width=\textwidth]{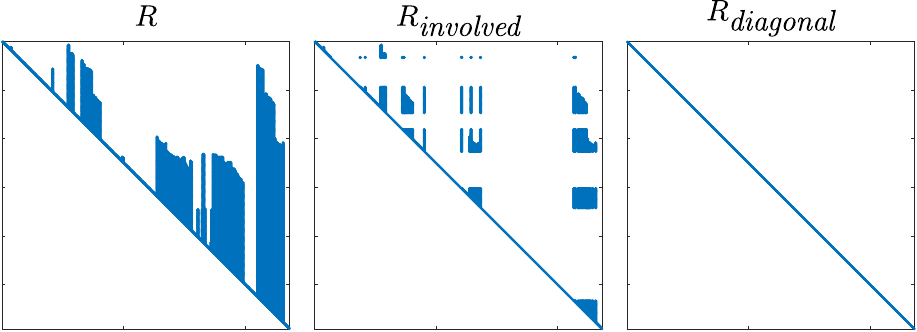}
  \caption{A square root matrix (taken from our experimental evaluation) and its sparse approximations generated with Algorithm~\ref{alg:sparsification}, for different variable selections $\mathcal{S}$. On the left -- the original matrix; in the center -- the matrix after partial sparsification, of only the uninvolved variables (here, about half of the variables); on the right -- the matrix after full sparsification. The matrices on the left and in the center are guaranteed to be action consistent. Full sparsification results in a convenient diagonal approximation of the information. For all degrees of sparsification, the determinant of the matrix remains the same.}
  \label{fig:information-comparison}
\end{figure}

Algorithm~\ref{alg:sparsification} summarizes our suggested method for belief sparsification.
The algorithm may receive as input, and return as output, a belief represented using either the information matrix, or its square root. This scalable algorithm depends on a pre-selected subset $\mathcal{S}$ of state variables, and wisely removes elements which correspond to these variables from the matrix. Approximations of different degrees can be generated using different variable selections $\mathcal{S}$, as to be explained in Section~\ref{sec:dmuu-var-selection}. For a clear discussion, when $\mathcal{S}$ contains all the variables, we say this is a \emph{full sparsification}; using any other partial selection of variables is a \emph{partial sparsification}.
Fig.~\ref{fig:sparsification-matrices} contains a visual demonstration of the algorithm steps.
In the following section (Section~\ref{sec:prob-analysis}), we provide an extended probabilistic analysis of the algorithm, and explain how it can also be applied to general (non-Gaussian) beliefs; a visual demonstration of such application, where we represent the belief using a generic factor graph, is given in Figure~\ref{fig:sparsification-graphs}. An example of the the algorithm output is provided in Figure~\ref{fig:information-comparison}.

\begin{algorithm}
\caption{Scalable belief sparsification.}
\label{alg:sparsification} 
\SetKw{return}{return}
\SetKwBlock{inputs}{Inputs:}{}
\SetKwBlock{outputs}{Output:}{}

\footnotesize
\setstretch{1.18}

\nonl\inputs{
\nonl A belief $b = \mathcal{N}({\bm X}^*,\bm \Lambda^{-1})$, such that $\bm \Lambda =\bm R^T \bm R$ \\
\nonl A subset $\mathcal{S}$ of state variables to sparsify 
}
\nonl\outputs{
\nonl A sparsified belief $b_s \doteq \mathcal{N}({\bm X}^*,\bm \Lambda_s^{-1})$, such that $\bm \Lambda_s \doteq \bm R_s^T \bm R_s$
}
\BlankLine

%\If{ \label{alg:sparsification-line:if-ordered}}{
\tcp{reorder the state variables such that the variables in $\mathcal{S}$ are first in the state vector}
$ \bm P \,\leftarrow\,$ an appropriate (column) permutation matrix \\
\If{the algorithm input is $\bm \Lambda$ \label{alg:sparsification-line:if-info}}{
${\bm \Lambda}^p \,\leftarrow\ \bm P^T \bm \Lambda \bm P$  \label{alg:sparsification-line:permute}\\
${\bm R}^p \,\leftarrow\, \mathtt{chol}({\bm \Lambda}^p)$ \label{alg:sparsification-line:chol}
}
\ElseIf{the algorithm input is $\bm R$ \label{alg:sparsification-line:if-root}}{
${\bm R}^p \,\leftarrow\,$ modify ${\bm R}$ to convey appropriate variable reordering (see remark in the main text) \label{alg:sparsification-line:modify}\\
%$\bm \Lambda \,\leftarrow\, \bm R^T \bm R$ \label{alg:sparsification-line:form} \\
}
%}

${\bm R}^p_s \,\leftarrow\,$ zero off-diagonal elements from ${\bm R}^p$ in rows matching variables in $\mathcal{S}$ \label{alg:sparsification-line:sparsify} \DontPrintSemicolon\tcp*{sparsify ${\bm R}^p$}
${\bm R}_s \,\leftarrow\, \bm P \bm {\bm R}^p_s \bm P^T$ \label{alg:sparsification-line:permute-back} \DontPrintSemicolon\tcp*{return to the original variable order}
\If{the algorithm output is $\bm \Lambda$}{
$\bm \Lambda_s \,\leftarrow\, \bm R_s^T \bm R_s$ \label{alg:sparsification-line:reform} \DontPrintSemicolon\tcp*{reform the information matrix}
}
\end{algorithm}

Let us break down the algorithm steps:
first, we should check if the variables are ordered properly, i.e., such that the variables we wish to sparsify (variables in $\mathcal{S}$) appear first in the state. If not, we should reorder the variables accordingly. This requires appropriate modification of the input matrix.
%\draft{This permutation is required in order to optimize fill-in, and to properly decorrolate the sparsified variables from the unsparsified ones -- a requirement in the proof of Therorem~\ref{trm:uninvolved-is-ac}, which is to appear in the next section, and which provides certain guarantees for the approximated belief. }
If the algorithm input is the symmetric matrix $\bm \Lambda$ (line~\ref{alg:sparsification-line:if-info}), we shall simply permute its rows and columns by calculating the product $\bm P^T \bm \Lambda \bm P$ of the information matrix with an appropriate (column) permutation matrix~$\bm P$.
After this permutation, we can derive ${\bm R}^p$, the square root matrix of the permuted information matrix, using the Cholesky decomposition (line~\ref{alg:sparsification-line:chol}).
If the algorithm input is the matrix $\bm R$ (line~\ref{alg:sparsification-line:if-root}), the task of variable reordering is not trivial, as trying to modify $\bm R$ by permuting its rows and columns would break its triangular shape. Instead, this task (typically) requires re-factorization of $\bm \Lambda$ under the new variable order. 

\begin{remark}
In our follow-up work \citep{Elimelech21ral}, we provide an efficient modification algorithm for $\bm R$, which is intended for the task of variable reordering, and can spare the matrix re-factorization; we can use this algorithm to efficiently derive ${\bm R}^p$ (line~\ref{alg:sparsification-line:modify}).
\end{remark}

\pagebreak

If no reordering is required, and the algorithm input is $\bm \Lambda$, we may directly calculate the Cholesky decomposition (line~\ref{alg:sparsification-line:chol}); if no reordering is required, and the input is $\bm R$, we may skip directly to line~\ref{alg:sparsification-line:sparsify}. Specifically, when all of $\mathcal{S}$ is already at the beginning of the state, no reordering is needed. This situation particularly occurs when sparsifying \textit{all} the variables (i.e., full sparsification).
Next, in line~\ref{alg:sparsification-line:sparsify}, we zero off-diagonal elements in the permuted square root matrix~${\bm R}^p$, in rows corresponding to variables in $\mathcal{S}$, to yield the sparsified square root matrix~${\bm R}^p_s$. 

Since the prior belief should be updated according to the predicted hypotheses, the variable order in the sparsified information matrix (or its square root) must match the variable order in the collective Jacobians. Thus, we should reorder the variables back to their original order (line~\ref{alg:sparsification-line:permute-back}). Though, we notice that after the sparsification this permutation can be performed on the square root matrix \emph{directly}, without resorting to the information matrix, and without breaking its triangular shape, by calculating $ \bm P \bm R^p_s \bm P^T $ (note the reverse multiplication order). This claim is formalized in Corollary~\ref{cor:reordering-in-root} (and proved in Appendix~\ref{apndx:proofs}). 

\begin{corollary}
After sparsification of the square root matrix (line~\ref{alg:sparsification-line:sparsify} of Algorithm~\ref{alg:sparsification}), permutation of the variables back to their original order can be performed on the square root matrix directly, without breaking its triangular shape.
\label{cor:reordering-in-root}
\end{corollary}

Finally, we may return the sparsified belief, represented either with $\bm R_s$ or $\bm \Lambda_s$. In the latter case, this requires to (easily) reconstruct the sparsified information matrix from its sparsified root (line~\ref{alg:sparsification-line:reform}). After the sparsification, the value of the non-zero (NZ) entries in the sparsified information matrix may be different than the corresponding entries in the original matrix (including the diagonal), and new NZs may be added in compensation for the removed entries (factors).
Also, note that the permutation of variables back to their original order can potentially be skipped, by equivalently permuting the columns of all the candidate collective Jacobians, to match the altered order.

The derivation of $\bm R_p$ (in line~\ref{alg:sparsification-line:chol} or line~\ref{alg:sparsification-line:modify}), when conducted, is the costliest step of the algorithm, which defines its maximal computational complexity; we may recall that the complexity of the Cholesky decomposition is $O(n^3)$, at worst, where $n$ is the state size \citep{Hammerlin12book}. 
In comparison, the computational cost of the remaining steps, i.e., matrix permutation (lines~\ref{alg:sparsification-line:permute} and \ref{alg:sparsification-line:permute-back}), removal of matrix elements (line~\ref{alg:sparsification-line:sparsify}), and reconstruction of the information matrix (line~\ref{alg:sparsification-line:reform}), is usually minor.
Still, it should be noted that depending on the configuration, many of the steps are often not necessary. For example, as mentioned, when the input matrix is already in the desired order, the permutations can be skipped; this is specifically correct in full sparsification. In that case, if given the square root matrix as input, the algorithm holds an almost negligible complexity -- we only need to extract the matrix' diagonal. Also, in full sparsification, the sparsified information matrix, if required, can be reconstructed from its root in linear complexity, as both ${\bm R}_s$ and ${\bm \Lambda}_s$ are diagonal.

Nonetheless, we remind that the approach is meant to overall reduce the decision making time, as the time spent on performing the sparsification (performed once) is lower than the time saved in performing (the multiple) belief updates.
For example, since full sparsification leads to a diagonal approximation (information or its root), considering the collective Jacobians are sparse, belief updates can be performed with an almost linear complexity. %Thus, this method allows us to enjoy the benefits of a high quality information measure (entropy), at the computational cost of a simple information measure (such as trace), while experiencing minimal loss, as to be seen.
Also, since the cost of sparsification does not depend on the number of candidates or hypotheses, as this number grows, the relative "investment" in calculating the sparsification becomes less significant.

\subsubsection{Probabilistic analysis \label{sec:prob-analysis}}
Let us analyze the suggested sparsification algorithm from a wider perspective, using probabilistic graphical models.

As explained, the belief~$b$ (\ref{eq:belief-factors}) is constructed as a product of factors -- probabilistic constraints between variables, e.g., those induced by observations or constraints between poses. A belief can be graphically represented with a factor graph -- where variable nodes are connected with edges to the factor nodes in which they are involved. In Fig.~\ref{fig:sparsification-graphs}a, we can see an exemplary factor graph, which represents a belief $b$ with six variables and eight factors:
\begin{multline}
\label{eq:example-belief}
b(\bm X) \propto \\ f_{x_1}\cdot f_{x_1 l_1}\cdot f_{x_1 l_2}\cdot f_{x_1 x_2}\cdot f_{x_2 l_1}\cdot f_{x_2 l_2}\cdot f_{x_2 x_3}\cdot f_{x_3 l_3},
\end{multline}
where the state $\bm X \doteq [ x_1,l_1,l_2,x_2,x_3, l_3 ]^T$ contains three poses and three landmarks, and $f_{ij}$ is a factor between $i$ and $j$.
As explained, in the linear(ized) Gaussian system, the belief~$b$ is described with the information matrix~$\bm \Lambda$, as shown in Fig.~\ref{fig:sparsification-matrices}a. Off-diagonal non-zero entries in the information matrix $\bm \Lambda$ indicate the existence of factors between the corresponding variables.

The belief~$b$ can be factorized to a product of conditional probability distributions, in a process known as "variable elimination" \citep[see][]{Davis06book}:
\begin{equation}
b \propto \prod_{i=1}^{n-1} \prob({\bm X}_i|d({\bm X}_i))\cdot\prob({\bm X}_n),
\end{equation}
where $d({\bm X}_i)$ denotes the set of variables ${\bm X}_i$ is conditionally dependent on -- a subset of the variables which follow ${\bm X}_i$ according to the variable (elimination) order. 
Practically, fixing the variable order in the state sets the decomposition of the belief. Thus, according to Algorithm~\ref{alg:sparsification}, we begin the sparsification process by reordering the state variables, such that all variables in $\mathcal{S}$ appear first in the state. This step requires us to permute the information matrix accordingly (as shown in Fig.~\ref{fig:sparsification-matrices}b); here, we chose $\mathcal{S} = \{ x_1,l_2,x_2 \}$. Note that variables can be conditionally dependent even if there is no factor between them.
By starting the elimination with the variables in $\mathcal{S}$, we force conditional separation of the variables for sparsification and the remaining variables, i.e.,
\begin{equation}
b \propto \prob(\mathcal{S} \mid \neg\mathcal{S}) \cdot \prob(\neg\mathcal{S}).
\end{equation}
This means that the no variable in $ \neg\mathcal{S}$ is conditionally dependent on a variable in $\mathcal{S}$.

The factorization of the belief to a product of conditional probabilities can be graphically represented with a Bayesian network ("Bayes net"), as shown in Fig.~\ref{fig:sparsification-graphs}c. In this directed graph, the existence of an edge from node $i$ to $j$ indicates that $i \in d(j)$. 
As established by \cite{Dellaert06ijrr}, this factorization is equivalent to the factorization of the (permuted) information matrix~${\bm \Lambda}^p$ to its upper triangular square root~${\bm R}^p$ (Fig.~\ref{fig:sparsification-matrices}c).
The conditional probability distribution of the $i$-th variable corresponds to the respective row of ${\bm R}^p$.
Off-diagonal entries in that row represent the conditional dependencies: if the off diagonal entry ${\bm R}^p_{ij}$ is non-zero, then ${\bm X}_j$ is in $d({\bm X}_i)$, and ${\bm X}_j$ is a parent of ${\bm X}_i$ in the Bayes net; specifically, if all elements on the $i$-th row, besides the diagonal entry, are zero, then ${\bm X}_i$ is not conditionally dependent on any variable (according to the elimination order), and has no parents in the Bayes net. \linebreak For more details, see \cite{Dellaert17foundations}.

According to the next step in the algorithm, we shall now zero off-diagonal entries in ${\bm R}^p$, in the rows which correspond to variables in $\mathcal{S}$ (Fig.~\ref{fig:sparsification-matrices}d); equivalently, this process can be seen as removing edges from the Bayes net (Fig.~\ref{fig:sparsification-graphs}d). By removing all the off-diagonal entries from the $i$-th row, we replace the conditional probability distribution 
\begin{equation}
\prob({\bm X}_i|d({\bm X}_i)) = \mathcal{N}\left(\mu(d({\bm X}_i)), ({{\bm R}^p_{ii}}^T {\bm R}^p_{ii})^{-1}\right)
\end{equation}
with an independent probability distribution over ${\bm X}_i$,
\begin{equation}
\prob_s({\bm X}_i) \doteq \mathcal{N}\left(\mu_i, ({{\bm R}^p_{ii}}^T {\bm R}^p_{ii})^{-1}\right).
\end{equation}
Essentially, we fix the mean of $\bm X_i$ to a constant value, which is no longer dependent on other variables. We, of course, would like to preserve the mean of the overall belief, and therefore shall select $\mu_i = \bm X^*_i$. It should be mentioned that this probability distribution is \emph{not} the marginal distribution over ${\bm X}_i$, which is given as $\mathcal{N}\left(\bm X^*_i, \Sigma_{ii}\right)$.

The sparisified belief is thus given as the product
\begin{equation}
b_s \propto \prod_{x \in \mathcal{S}} \prob_s(x) \cdot \prob(\neg\mathcal{S}).
\end{equation}
The chosen elimination order makes sure that the inner dependencies among the non-sparsified variables remain exact.
Notably, the suggested sparsification is performed by manipulating the square root matrix, which is equivalent to manipulating the Bayes net. In contrast, traditional belief sparsification methods (as we reviewed) perform sparsification on $\bm \Lambda$ directly, or equivalently, the factor graph. Still, we would like to understand what the factor-decomposition, which corresponds to the sparsified~belief,~is. 
Let us look again at the exemplary belief, given in (\ref{eq:example-belief}).
We begin its factorization (after the initial reordering) by eliminating the variables in $\mathcal{S}$ (in order). First, $x_1$: 
\begin{multline}
b \propto \prob(x_1\mid x_2,l_1,l_2)\cdot \hfill\\\hfill f'_{x_2 l_1 l_2}\cdot f_{x_2 l_1}\cdot f_{x_2 l_2}\cdot f_{x_2 x_3}\cdot f_{x_3 l_3}.
\end{multline}
Then, $l_2$: 
\begin{multline}
b \propto \prob(x_1\mid x_2,l_1,l_2)\cdot \prob(l_2\mid x_2,l_1)\cdot \hfill\\\hfill f'_{x_2,l_1}\cdot f_{x_2 l_1}\cdot f_{x_2 x_3}\cdot f_{x_3 l_3}.
\end{multline}
Finally, $x_2$: 
\begin{multline}
\label{eq:partial-elimination}
b \propto \prob(x_1\mid x_2,l_1,l_2)\cdot \prob(l_2\mid x_2,l_1)\cdot \prob(x_2\mid l_1, x_3) \cdot \hfill\\\hfill f'_{x_3 l_1}\cdot f_{x_3 l_3}.
\end{multline}
This partial elimination is visualized in Fig.~\ref{fig:sparsification-graphs}b.
As we can see, after elimination of variables, new "marginal" factors ($f'_{x_2 l_1 l_2}, f'_{x_2,l_1}, f'_{x_3 l_1}$) may be introduced to the belief, representing new links among the non-eliminated variables; in our case, after eliminating all the sparsified variables, one marginal factor still remains: $f'_{x_3 l_1}$.

According to the previous analysis, in the sparsification, each of the conditional distributions on the sparsified variables is replaced with an independent distribution. These are, in fact, unitary factors over the variables; here, we mark those as $f''_{x_1},\, f''_{l_1},\, f''_{x_2}$. 
The sparsified belief can thus be given as a product of these unitary factors on the sparsified variables, the marginal factors introduced after eliminating these variables, and the remaining non-eliminated factors (here, $f_{x_3 l_3}$). Overall, in our example, this product is:
\begin{equation}
\label{eq:sparse-belief-factors}
b_s \propto f''_{x_1}\cdot f''_{l_1}\cdot f''_{x_2}\cdot f'_{x_3 l_1}\cdot f_{x_3 l_3}
\end{equation}
The factor graph matching this belief is shown in Fig.~\ref{fig:sparsification-graphs}e. \linebreak It is clear that the sparsification does not affect the elimination of the remaining variables (variables in $\neg\mathcal{S}$). Continuing the elimination process from either $b$~(\ref{eq:partial-elimination}) or $b_s$~(\ref{eq:sparse-belief-factors}) would result in the same distribution $\prob(\neg\mathcal{S})$.

To complete the analysis, we shall note that this sparsification method does not change the diagonal entries in the information root matrix, and, thus, the determinants of $\bm \Lambda$ and $\bm \Lambda_s$ remain the same:
\begin{multline}
\label{eq:sparse-det-eq}
\abs{{\bm \Lambda}} = \abs{{\bm \Lambda}^p} = \abs{{{\bm R}^p}^T {\bm R}^p} = \hfill\\\hfill \abs{{\bm R}^p}^2 = \prod^n_i ({\bm R}^p_{ii})^2 = \abs{{\bm R}^p_s}^2 \hfill\\\hfill= \abs{{{\bm R}^p_s}^T {\bm R}^p_s} = \abs{{\bm \Lambda}^p_s} = \abs{{\bm \Lambda}_s}.
\end{multline}
Hence, the sparsification method preserves the overall entropy of the belief (as defined in (\ref{eq:entropy})), no matter which variables are sparsified. This is usually not guaranteed in the aforementioned traditional sparsification methods. Still, when incorporating new factors in the future, divergence in entropy between the original and sparsified beliefs (i.e., simplification offset) might indeed happen. 
This offset depends on the variables selected for sparsification, and can even be zero, as we shall discuss next. Since the sparsified variables become independent, if we wish to update our estimation after applying new actions, or after acquiring a new observation of an existing variable (i.e., loop closure), information would no longer propagate from a sparsified variable to another variable, or vice-versa, unless they are observed together. 
Though, notably, unlike simply marginalizing the sparsified variables out of state, as done in filtering, they can still be updated in the future.

\subsection{Optimality guarantees}
\subsubsection{Variable selection and pre-solution guarantees \label{sec:dmuu-var-selection}}
Next, we shall present the conclusions of our symbolic analysis of the suggested simplification method (as explained in Section~\ref{sec:guarantees-general}). In this evaluation, we utilized our knowledge on the decision problem formulation, and on Algorithm~\ref{alg:sparsification}, in order to derive general guarantees for the simplification loss. More specifically, we shall explain which variables should be sparsified, such that the effect on the objective value for each candidate action (i.e., the simplification offset) is minimal.

\begin{figure}[b]\center
  \includegraphics[trim= 0 -10 0 0, width=\textwidth]{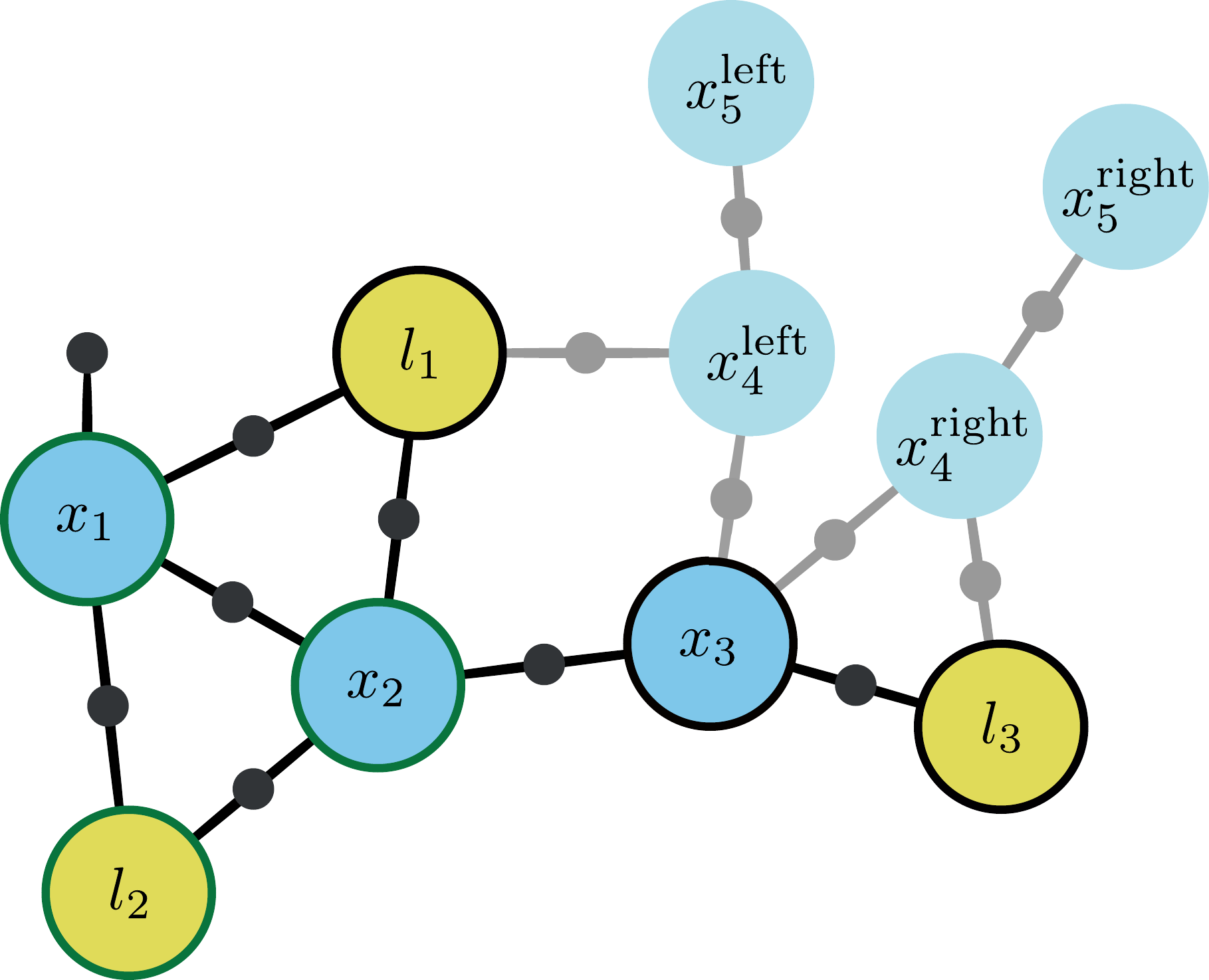}
  \caption{A factor graph representing the belief of an agent in an exemplary full-SLAM scenario. The current (prior) state consists of three poses $x_1,x_2,x_3$ (blue nodes), and the position of three landmarks $l_1,l_2,l_3$ (yellow nodes), which were previously observed. Factors (black nodes) between poses mark motion constraints, and factors between a pose and a landmark mark observation constraints. At time of planning, the agent is at pose $x_3$, and wishes to infer which of the candidate paths $\mathcal{U}=\{\text{left},\,\text{right}\}$ is the optimal one. If taking the right path, the agent predicts augmenting its state with two new poses $x^\text{right}_4, x^\text{right}_5$, with motion constraints connecting them to the current pose; based on its current state estimation, it also predicts observing landmark $l_3$ from $x^\text{right}_4$ (i.e., adding an observation constraint between $l_3$ and the new pose). The variables (from the prior state) involved with this action are those directly connected to any of the predicted new factors -- $x_3,l_3$.
If taking the left path, the agent predicts augmenting its state with two new poses $x^\text{left}_4,x^\text{left}_5$, and observing landmark $l_1$  from $x^\text{left}_4$. The variables involved with this action are $x_3,l_1$.
The involved variables (in any of the actions) are marked with black outline.
Note that $x_1,x_2,l_2$ are never involved; these are marked with dark green outline.
Theorem~\ref{trm:uninvolved-is-ac} suggests that the uninvolved variables can be sparsified from the prior belief (via Algorithm~\ref{alg:sparsification}), while maintaining action consistency.}
  \label{fig:factor-graph}
\end{figure}

Considering a specific action, a state variable is \emph{involved} if applying the action adds a constraint (factor) on it; \linebreak i.e., if $g$ or $h$, which define the relevant transition and observation models (which are defined in (\ref{eq:model-trans}) and (\ref{eq:model-obs})), are affected by this variable. Practically, in the collective Jacobian of an action, each of the columns corresponds to a state variable, and every row represents a constraint; a variable is involved if at least one of the entries in its matching column is non-zero; \emph{uninvolved} variables correspond to columns of zeros. For example, in a navigation scenario, the landmarks we predict to observe by taking the action (along with the current pose) are involved; variables referring to landmarks from the past, which we do not predict to observe, are uninvolved. 
An illustration of this example is given in Fig.~\ref{fig:factor-graph}.

We emphasize that since this is a planning problem, the collective Jacobians, the objective values, and the involved variables are \emph{determined} based on our prediction for the outcome of each action. Further, these components can only be based on our current belief, and not the ground truth, as it is unknown.
Thus, although a landmark we identified as uninvolved, might be observed when applying the action (e.g., if the initial belief was distant from the ground truth), this is not a concern in the planning context. As explained, in our formulation, the objective function (\ref{eq:objective}) relies on the "most likely" observation. In other words, we consider only the single "most likely" outcome for each action. Theoretically, we can consider multiple probabilistic outcomes for each action, each determining its own set of involved variables; as mentioned, this generalized discussion is brought by \cite{Elimelech21thesis}.

We claim that for any given action, sparsifying the uninvolved variables from the prior belief $b$, before computing the posterior belief, would not affect the posterior entropy (which defines our objective function $V$). Hence, for a set of candidate actions $\mathcal{U}$, we can sparsify from the prior belief all variables which are uninvolved in \emph{any} of the actions, and use this sparsified belief $b_s$ to compute the objective function, without affecting its values. Specifically, this means that the simplification offset is zero, and that this sparsified belief is action consistent with the original one: $ b \simeq b_s $. This claim is formally expressed in Theorem~\ref{trm:uninvolved-is-ac}. A~proof for this claim is given in Appendix~\ref{apndx:proofs}.

\begin{theorem}
Consider a decision problem $\problem \doteq \left(b,\mathcal{U},V\right)$,\linebreak where $b$  is a (Gaussian) initial belief, and $V$ is the objective function from (\ref{eq:objective}).
Considering a set $\mathcal{S}$ of state variables, which are uninvolved in any of candidates in $\mathcal{U}$, \linebreak Algorithm~\ref{alg:sparsification} returns a belief $b_s$, such that $\,\Delta(\problem,\problem_s) = 0,\,$ where $\,\problem_s \doteq \left(b_s,\mathcal{U},V\right)$.
\label{trm:uninvolved-is-ac}
\end{theorem}

In principle, only a single sparsification process is conducted for each decision problem (i.e.,~planning session), regardless of the number of candidate actions. Selecting variables which are uninvolved in any of the candidate actions allows to keep action consistency considering the entire set of candidates. Still, it is possible to break the set of actions to several subsets of similar actions, and consider the uninvolved variables in each subset. For each subset we would create a custom prior approximation, and then select the best candidate in each of the subsets, before finding the overall best candidate among those. This can result in a more adapted sparsification for each subset. Yet, calculation of the sparsification itself has a cost, which needs to be considered when trying to achieve the best performance. Here we examine the most general case -- treating the set of actions as a whole. 
 
\begin{remark}
We note that if we consider (1) sparsification of only uninvolved variables; (2) the output of Algorithm~\ref{alg:sparsification} to be the square root matrix; and (3) no requirement to maintain the original variable order after the sparsification (by instead, reordering the collective Jacobians); then, there is no need to actually zero entries in the rows of the "sparsified" variables. The initial reordering is sufficient to make sure that these rows would not be updated when (incrementally) incorporating new constraints. An in-depth look at this variation was examined in our follow-up work \citep[see][]{Elimelech19isrr}.
\end{remark}

We proved that sparsifying uninvolved variables does not affect the objective function values, and, therefore, they should always be included in the set $\mathcal{S}$ of variables for sparsification. It is possible to sparsify also involved variables, but then "zero offset" and action consistency are not guaranteed. Intuitively, selecting more involved variables to $\mathcal{S}$ results in a sparser approximation, but potentially a larger divergence from the original objective values. In Appendix~\ref{apndx:pre-eval-guarantees}, we show that under additional restrictions, we can symbolically derive offset (and loss) bounds also when sparsifying involved variables; these bounds are only applicable for "rank 1" updates, i.e., when the collective Jacobians are limited to a single row.

\subsubsection{Post-solution guarantees \label{sec:dmuu-guarantees}}
For a more general scenario, when sparsifying involved variables, and with actions possibly having multi-row collective Jacobians, we can try to bound the loss by performing post-solution analysis, as discussed in Section~\ref{sec:guarantees-general}. Unlike before, such guarantees are derived \emph{after} solving the simplified problem (but before applying the selected action). 
As explained, we can utilize the calculated (simplified) objective values, and domain-specific lower and upper bounds of the objective function ($\mathcal{LB},\mathcal{UB}$, respectively), to yield offset bounds (\ref{eq:offset-bound-post}); from these offset bounds, we can then easily derive loss bounds (\ref{eq:loss-bound-post-post}).

As our decision problem domain relies on beliefs, which, as we saw, can be represented with a (factor) graph, we can potentially exploit topological aspects to derive the desired objective bounds. 
% Surely, we aspire to select the most cost-effective set $\mathcal{S}$, that gives the highest degree of sparsification under the given budget; yet, finding the optimal set $\mathcal{S}$ is by itself an optimization problem. This can be done by scalably enlarging $\mathcal{S}$ according to some heuristic, until the bound no longer guarantees satisfying results. Other options are random selections until the criteria is met, genetic algorithms, etc. While a wise selection of $\mathcal{S}$ also bares a certain cost, it can prove itself profitable, especially for a large set of candidate actions, where the initial investment becomes less significant. Nonetheless, in the results to follow, we demonstrate that even when sparsifying all the variables, the quality of solution is still well preserved.
For example, we can utilize conclusions from a recent work by~\cite{Kitanov19arxiv}, which extends a previous work by \cite{Khosoussi18ijrr}. There, the following bounds on the information gain were proved, for when the corresponding factor graph contains only the agent's poses, and each pose consists of the position and the orientation of the agent (i.e., pose-SLAM): 
%\begin{fleqn}
\begin{empheq}[box=\fbox]{align}
&\mathcal{LB}_\text{top}\left\{ V(b,u) \right\} \doteq 3\cdot\ln t(b,u) + \mu + \entropy(b), \label{eq:lb-topological} \\% {n-1 \over2}\left[ \ln\abs{\Omega_\omega} - k\cdot\ln(2\pi e) \right],
&\mathcal{UB}_\text{top}\left\{ V(b,u) \right\} \doteq \nonumber\\
& \hspace{1cm} \mathcal{LB}_\text{top}\left\{ V(b,u) \right\} + \sum_{i=2}^n \ln( d_i + \Psi) - \ln\abs{\tilde {\bm L}}, \label{eq:ub-topological}
\end{empheq}
where $t(b,u)$ stands for the number of spanning trees in the factor graph of the posterior belief ($b$ after applying $u$); $n$ marks the graph size; $\tilde {\bm L}$ is the reduced Laplacian matrix of the graph; and $d_i$'s are the node degrees corresponding to $\tilde {\bm L}$. They also assume that the factors between the poses are described with a constant diagonal noise covariance; $\mu$~and~$\Psi$ are constants which depend on this noise model, and the posterior graph size (i.e., the length of the action sequence). In their demonstration, they show that when the ratio between the angular variance and the position variance is small, these bounds are empirically tight. This case can happen, for example, when a navigation agent is equipped with a compass, which reduces the angular noise. For a detailed derivation of these bounds please refer to \cite{Kitanov19arxiv}.

For different problem domains, it is possible to use various other objective bounds in a similar manner. For example, in Appendix~\ref{apndx:post-eval-guarantees}, we present additional bounds, which exploit known determinant inequalities. These make no assumptions on the state structure, and are potentially useful when the matrix $\bm \Lambda$ is diagonally dominant.

\section{Experimental results}
\graphicspath{ {./images/scenario/} }

\newcommand{\scenarioFigure}[2] {
\graphicspath{ {./images/scenario/#1/} }
\begin{figure}[H]
    \begin{subfigure}[t]{\textwidth}
	  \includegraphics[width=\textwidth]{screenshot}
	  \caption{A screenshot of the scenario, which includes: the map estimation (blue occupancy grid); the current estimated position (yellow arrow-head) and goal (yellow circle); the trajectory taken up to that point (thin green line); the candidate trajectories from the current position to the goal (thick lines in various colors); and the selected trajectory (highlighted in bright green). \\ $ $}
    \end{subfigure}
    \begin{subfigure}[t]{0.60\textwidth}
		\includegraphics[width=\textwidth]{fig_rewards-eps-converted-to.pdf}
        \caption{Objective function comparison. \\ $ $}
    \end{subfigure}
    \begin{subfigure}[t]{0.39\textwidth}
		\includegraphics[width=\textwidth]{fig_timeY-eps-converted-to.pdf}
        \caption{Run-time. \\ $ $}
    \end{subfigure}
    \begin{subfigure}[t]{0.99\textwidth}
		\includegraphics[width=\textwidth]{fig_information_root-eps-converted-to.pdf}
        \caption{Original prior information root matrix and its sparse approximations.}
    \end{subfigure}
    \begin{subfigure}[t]{\textwidth}
		\includegraphics[width=\textwidth,center]{fig_actions-eps-converted-to.pdf}
        \caption{Collective Jacobians of the candidate trajectories.}
    \end{subfigure}
	  \caption{Results summary for planning session \##2.}
	  \label{fig:scenario-#2} 
\end{figure}
%\addtocounter{figure}{-1}
}

\newcommand{\scenarioFigureCompact}[2] {
%\begin{framed}
\graphicspath{ {./images/scenario/#1/} }
\begin{figure}[H]
	\begin{subfigure}[t]{\textwidth}
        \includegraphics[width=\textwidth]{screenshot}
        \caption{The scenario. \\ $ $}
	\end{subfigure}
    \begin{subfigure}[t]{0.60\textwidth}
        \includegraphics[width=\textwidth]{fig_rewards-eps-converted-to.pdf}
        \caption{Objective function comparison.}
    \end{subfigure}
    \begin{subfigure}[t]{0.39\textwidth}
        \includegraphics[width=\textwidth]{fig_timeY-eps-converted-to.pdf}
        \caption{Run-time.}
    \end{subfigure}
   \caption{Results summary for planning session \##2}
    \label{fig:scenario-#2}
\end{figure}
}

\subsection{The scenario}
\begin{figure}[H]
	\includegraphics[width=0.49\textwidth]{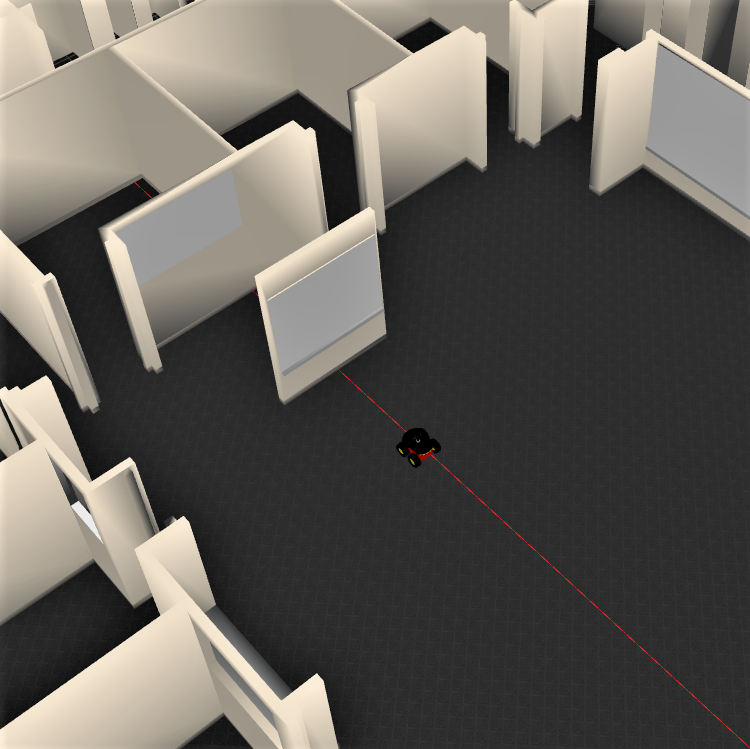}
	\includegraphics[width=0.49\textwidth]{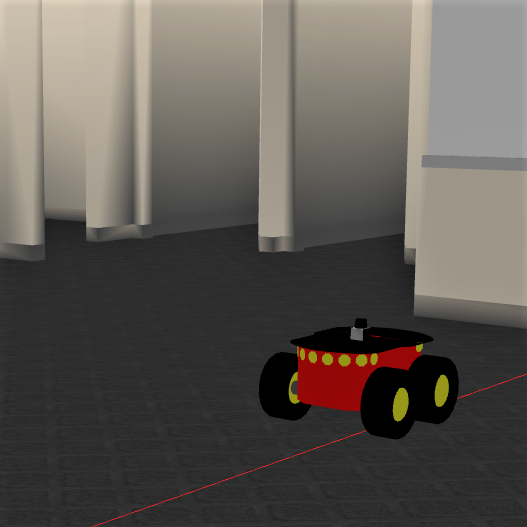}
	\caption{A Pioneer 3-AT robot in the simulated indoor environment. The robot is equipped with a lidar sensor, Hokuyo UST-10LX, as visible on top of it.}
	\label{fig:scenario-robot}
\end{figure}

To demonstrate the advantages of the approach, we applied it to the solution of a highly realistic active-SLAM problem. In this scenario, a robotic agent navigates through a list of goals in an unknown indoor environment. We used the Gazebo simulation engine \citep{Koenig04iros} to simulate the environment and the robot -- Pioneer 3-AT, which is a standard ground robot used in academic research worldwide. The robot is equipped with a lidar sensor, Hokuyo UST-10LX. These components can be seen in Fig.~\ref{fig:scenario-robot}. Despite examining a 2D navigation scenario, our method does not impose any restrictions on the pose size nor on the state~structure.

We used the pose-SLAM paradigm, meaning, the agent's state $\bm X_k \doteq (x^T_0,\dots,x^T_k)^T$ consists only of poses kept along its entire trajectory. Each of these poses consists of three variables, representing the position and orientation. Our approach is highly relevant in this case, in which the state size grows quickly as the navigation progresses, making the planning more computationally-challenging.
The belief over the state is represented as a factor graph, and implemented using the GTSAM C++ library \citep{Dellaert12tr}. When adding a new pose to the graph, the sensor scans the environment in a range of 30 meters, and provides a point-cloud of it. This point-cloud is then matched to scans taken in previous poses using ICP matching \citep{Besl92pami}. If a match is found, a loop-closure factor (constraint) is added between these poses. To keep the computation cost of the scan matching feasible, and to avoid creating redundant constraints, we make sure to compare the current pose only to key poses within a certain range of (estimated) distances from it. Transition (motion) constraints are also created between every two consecutive poses. Both the observation and motion contain some Gaussian noise, which matches the real hardware's specs. Robot Operating System (ROS) is used to run and coordinate the system components -- state inference, decision making, sensing, etc.

\begin{figure}[b]
	\includegraphics[width=\textwidth]{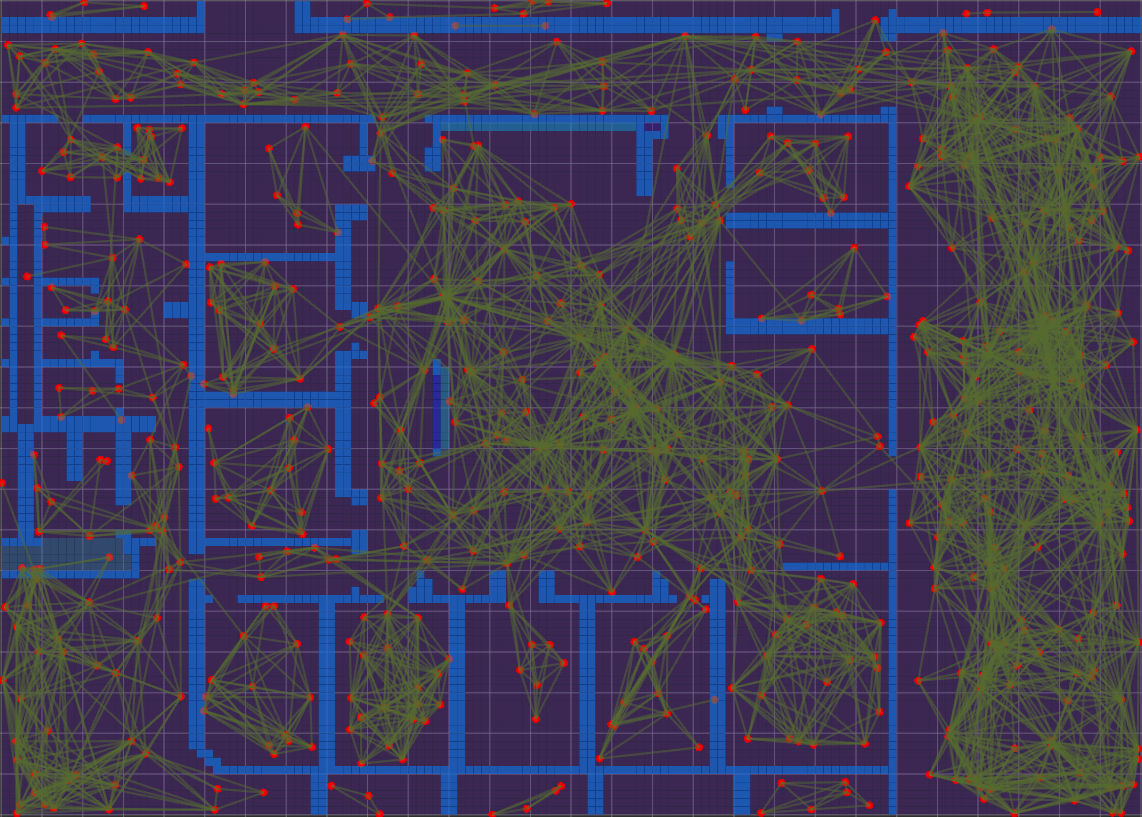}
	\caption{The entire indoor environment from a top view. Walls are colored in light blue. The PRM graph, from which trajectories are built, is colored in red and green. Each square on the map represents a 1m$\times$1m square in reality.}
	\label{fig:scenario-prm}
\end{figure}

The full indoor map is unknown to the robot, and it is incrementally approximated by it using the scans during the navigation. We do, however, rely on the full and exact map to produce collison-free candidate trajectories. We use the Probabilistic RoadMap (PRM) algorithm \citep{Kavraki96tra} to sample that map, and then use the K-diverse-paths algorithm \citep{Voss15icra} to build a set $\mathcal{U}$ of trajectories to the current goal. This usage of the map is irrelevant to the demonstration of our method; in our formulation, we consider the candidate actions are given. The complete indoor map is shown in Fig.~\ref{fig:scenario-prm}, with the sampled PRM graph on it. Each trajectory matches, of course, a certain control sequence, and is translated to a series of factors and constraints to be added to the prior factor graph. Loop closure constraints are added between poses in the new trajectory, and poses in the previously-executed trajectory, according to their estimated location (i.e., where we expect to add them when executing this trajectory). The corresponding collective Jacobians of the candidate trajectories are constructed as explained in Section~\ref{sec:bsp-def}. 

\begin{table*}[b]
\footnotesize
\caption{Numerical summary for all sessions. "Uninvolved var. ratio" represents the percentage of uninvolved variables in the prior state. "Run-time" represents the reduction in decision making time in the specified configuration, in comparison to the original problem. "Non zeros" represents the reduction in the number of non-zero entries in the prior square root matrix, after using the sparsification. "Sparsification time" represents the cost of this one-time calculation, out of the entire problem run-time.}
\label{tbl:scenario-runtime}
\begin{tabularx}{\textwidth}{|l?X?X|X|X|X?X|X|X|}
\hline \multirow{2}{*}{\textbf{Session}} &\multirow{2}{*}{\textbf{Prior Size} }  &\multicolumn{4}{c?}{\small \textbf{$\mathcal{P}_\textit{involved}$}}  &\multicolumn{3}{c|}{\textbf{\small $\mathcal{P}_\textit{diagonal}$}} \\
\cline{3-9}
& & {Uninvolved \newline var. ratio} 
& {Run-time} 
& {Sparsifica- tion time} 
& {Non zeros} 
& {Run-time} 
& {Sparsifica- tion time} 
& {Non zeros}
\\\Xhline{3\arrayrulewidth}
1  & 567 & 46\% & -23\% & 3\% & -76\% & -55\% & 1\% & -97\%
\\\hline
2  & 762 & 74\% & -34\% & 4\% & -77\% & -67\% & 1\% & -98\%
\\\hline
3  & 1182 & 60\% & -66\% & 1\% & -83\% & -85\% & 1\% & -99\%
\\\hline
4  & 1269 & 69\% & -70\% & 2\% & -86\% & -86\% & 2\% & -99\%
\\\hline
5  & 1341 & 65\% & -67\% & 2\% & -84\% & -82\% & 2\% & -99\%
\\\hline
6  & 1392 & 44\% & -52\% & $<$1\% & -61\% & -80\% & $<$1\% & -99\%
\\\hline
\end{tabularx}
\end{table*}

Since all trajectories lead to the goal, we only wish to optimize the "safety" of taking the path. Meaning, keeping the uncertainty of the state low, by preferring a more informative trajectory. We use the aforementioned objective function $V$ (from (\ref{eq:objective-root})) to compare between candidates. Under the "maximum likelihood" assumption, our method is only relevant to the computation of this information-theoretic measure, so for a more convenient discussion, we do not consider other objectives, such as the length of the trajectory. 

To cover its list of goals, the robot executes several planning sessions. In each session, the robot is provided with one goal, generates a set of candidate trajectories $\mathcal{U}$ to it, and selects the best candidate by solving a decision problem. The robot completes executing the entire selected trajectory before starting a new planning session to the next goal. To evaluate our method, in each planning session, we solved three decision problems, with each problem using another version of the initial belief. The robot's original initial belief accounts for the trajectory of poses executed up to that point (the entire inferred state). The other two versions are generated by sparsifying the original belief using Algorithm~\ref{alg:sparsification} -- one with partial sparsification, and one with full sparsification. Overall, in each session, the three configurations of the decision problem are as follows:
\begin{enumerate}
\item $\mathcal{P} = \left( b, \mathcal{U}, V \right)$ -- the original decision problem;
%\vspace{10pt}
\item $\mathcal{P}_\textit{involved} = \left( b_\textit{involved}, \mathcal{U}, V \right)$ -- with sparsification of the uninvolved variables -- an action consistent problem. We remind again that uninvolved variables correspond to columns of zeros in the collective Jacobians of all candidate actions, as explained in Section~\ref{sec:dmuu-var-selection}.
%\vspace{10pt}
\item $\mathcal{P}_\textit{diagonal} = \left( b_\textit{diagonal}, \mathcal{U}, V \right)$ -- with sparsification of all variables, leading to a diagonal information matrix, but not necessarily action consistent.
\end{enumerate}

For each configuration, we measured the objective function calculation time for each candidate action, along with the one-time calculation of the sparsification itself for the latter two. On the whole, in each planning session, we measure the total decision making time for each of the three configurations. For a fair comparison of the problems, the objective function calculation was detached from the factor graph-based implementation of the belief. From GTSAM, we extracted the square root matrix of the initial belief, and the collective Jacobians corresponding to (the factors added by) each candidate trajectory. Then, using Algorithm~\ref{alg:sparsification}, we created the two additional versions of the prior matrix, as detailed before. For each of the three decision problems, i.e., using each version of the prior square root matrix, we calculated the corresponding posterior square root matrix (via QR update); as explained in Section~\ref{sec:root-form}, we could then easily extract the determinant of these triangular matrices, to calculate the objective values.

%\pagebreak

At the end of each session, we applied the action selected by configuration 1. Of course, in a real application we would only solve the problem using a single configuration; here we present a comparison of the results for different configurations. We also did not invest in smart selection of variables for sparsification, as even full sparsification achieved very accurate results.

\subsection{Results}
In the following section we present and analyze the results from a sequence of six planning sessions. Of course, these sessions took place after the robot had already executed a certain trajectory in the environment, in order to build a state in a substantial size, and a map; if the prior state is empty, examining its sparsification is vain. Figs.~\ref{fig:scenario-1}-\ref{fig:scenario-6} showcase a summary of each of the planning sessions, and contain several components:

(a) A screenshot of the scenario, which includes: the map estimation (blue occupancy grid); the current estimated position (yellow arrow-head) and goal (yellow circle); the trajectory taken up to that point (thin green line); the candidate trajectories from the current position to the goal (thick lines in various colors); and the selected trajectory (highlighted in bright green).

(b) A comparison of the objective function values of the candidate actions (i.e., trajectories), considering each of the versions of the initial belief: $\mathcal{P}$ with the original belief in red; $\mathcal{P}_\textit{involved}$ with sparsification of the uninvolved variables in blue; and $\mathcal{P}_\textit{diagonal}$ with sparsification of all the variables in green. For scale, the comparison also contains the prior differential entropy, before applying any action. This "prior value" is not affected by the sparsification, and is the same for the three configurations (see~(\ref{eq:sparse-det-eq})).

(c) A comparison of the the solution time for the three decision problems. Again, $\mathcal{P}$ in red, $\mathcal{P}_\textit{involved}$ in blue, and $\mathcal{P}_\textit{diagonal}$ in green. The highlighted parts of the blue and green bars mark the cost of the sparsification itself out of the total solution time.

(d) A comparison of the three versions of the triangular square root matrix. The figures indicate non-zero entries in each matrix, i.e., their sparsity pattern.

(e) The sparsity pattern of the collective Jacobians of the examined trajectories. Again, uninvolved variables are identified by having columns of zero in all the Jacobians.

For the first and last sessions we provide an in-depth inspection, including all the components. Since the structure of the belief and Jacobians in all the sessions is similar, for the intermediate sessions we only present a summarized version, with components (a)-(c). The square root matrix and its approximations, given previously in Fig.~\ref{fig:information-comparison}, are extracted from the third session. Additionally, the numerical data shown in the figures is summarized in Table~\ref{tbl:scenario-runtime}. Further data regarding the loss is later given in Table~\ref{tbl:scenario-loss}.

\subsubsection{Efficiency}
As expected, the sparsification leads to a significant reduction in decision making time. The simplified problem $\mathcal{P}_\textit{diagonal}$ consistently achieves the best performance, followed by $\mathcal{P}_\textit{involved}$, while both are vastly more efficient than the original problem $\mathcal{P}$. Surely, a higher degree of sparsification ($\mathcal{S}$ containing more variables) leads to a greater improvement in computation time.
As discussed in Section~\ref{sec:algo}, full sparsification of the square root matrix has a particularly low cost -- we only need to extract its diagonal.
From Table~\ref{tbl:scenario-runtime} and the run-time comparison bar diagrams, it is clear that the cost of a partial sparsification is also minor in relation to the entire decision making. In some of the diagrams, the highlighted section of the bar, which stands for the cost of the sparsification, is hardly visible. Also, since the sparsification cost does not depend on the number of candidate actions, the larger the set of actions is, the less significant the sparsifcation should become.

We see a correlation between the ratio of uninvolved variables and the reduction in run time with $P_{involved}$. Variables corresponding to the executed trajectory become involved when a loop closure factor is created between them and a candidate trajectory. Hence, the ratio of uninvolved variables represents the overlap of the candidate trajectories with the previously executed trajectory. In the first session, the executed trajectory is short, resulting in a relatively small state size, and sparse root matrix, since not many loop closures were formed. As the sessions progress, the prior matrix becomes larger and denser, due to new loop closures, as apparent in the sixth session. 

In principle, we also notice a correlation between the state size and relative improvement in performance, for both sparsification configurations. Updating the square root factorization, in order to calculate the posterior determinant, has, at worst, cubical complexity in relation to the matrix size. An update to a variable at the beginning of the state (i.e., a loop closure) may force us to recalculate the entire factorization, baring this maximal computational cost. Sparsification of variables reduces the number of elements to update, and thus should be more beneficial when handling larger and denser beliefs.

\subsubsection{Accuracy}
Alongside the undeniable improvement in efficiency, we can also examine the quality of the selected action. According to Theorem~\ref{trm:uninvolved-is-ac}, not only $\mathcal{P}$ and $\mathcal{P}_\textit{involved}$ are action consistent, but they produce exactly the same objective values. Hence, solving $\mathcal{P}_\textit{involved}$ always leads to the optimal action selection, and induces no loss. $\mathcal{P}_\textit{diagonal}$ is not always action consistent with the original problem, and maintaining the same action selection is not guaranteed; however, it is evident from Figs.~\ref{fig:scenario-1}-\ref{fig:scenario-6} that even when sparsifying all the variables, the quality of solution is maintained. Not only does the graphs of $\mathcal{P}$ and $\mathcal{P}_\textit{diagonal}$ maintain a very similar trend, which practically leads to the same action selection, and zero loss, but also the difference (offset) between them is slim.
This is also evident by examining the Pearson rank correlation coefficient $\rho$ (which we mentioned in Section.~\ref{sec:analyzing-simp}) between the solutions of the original and simplified decision problems. A value of $\rho=1$ represents perfect correlation of the candidate rankings (i.e., action consistency), and $\rho=-1$ represents exactly opposite rankings. Clearly, the calculated values, presented in Table~\ref{tbl:scenario-loss}, indicate that $\mathcal{P}_\textit{diagonal}$ indeed resulted in an action consistent solution (or very close to it).
We emphasize again, that regardless of the selected action, the inference of the next state remains unchanged, as it is done on the original belief.

\pagebreak

\scenarioFigure{141201}{1} %1
\newpage
\scenarioFigureCompact{141956}{2} %2
%\scenarioFigureCompact{142741}
\scenarioFigureCompact{144327}{3} %3
\newpage
\scenarioFigureCompact{145355}{4} %4
\scenarioFigureCompact{150158}{5} %5
\newpage
\scenarioFigure{151629}{6} %6

\begin{table*}[t]
\footnotesize
\caption{The loss induced by the two simplified configurations, alongside the bounds on the loss (of the diagonal configuration), \linebreak for different noise models. The specified ratio for each bound represents the ratio between the angular variance and the position variance. No bound is calculated for the other configuration, since it is guaranteed to induce no loss. The loss and its bounds are brought as a percentage of the maximal approximated value in that session. Also shown is Pearson rank correlation coefficient $\rho$.}
\label{tbl:scenario-loss}
\begin{tabularx}{\textwidth}{|l?l|l?X|X?X|X|X|}
\hline
\textbf{Session}
& \textbf{$\rho(\mathcal{P},\mathcal{P}_\textit{involved})$}
& \textbf{$\rho(\mathcal{P},\mathcal{P}_\textit{diagonal})$}
& \textbf{$\textit{loss}(\mathcal{P},\mathcal{P}_\textit{involved})$} 
& \textbf{$\textit{loss}(\mathcal{P},\mathcal{P}_\textit{diagonal})$} 
& \textbf{$\textit{loss}(\mathcal{P},\mathcal{P}_\textit{diagonal})$ \newline bound -- 0.01:1}
& \textbf{$\textit{loss}(\mathcal{P},\mathcal{P}_\textit{diagonal})$ \newline bound -- 0.25:1} 
& \textbf{$\textit{loss}(\mathcal{P},\mathcal{P}_\textit{diagonal})$ \newline bound -- 0.85:1} 
\\\Xhline{3\arrayrulewidth}
1  & 1 & 0.99 & 0\% & 0\% & 2\% & 16\% & 46\%
\\\hline
2  & 1 & 1    & 0\% & 0\% & 2\% & 16\% & 47\%
\\\hline
3  & 1 & 1    & 0\% & 0\% & 1\% & 13\% & 39\%
\\\hline
4  & 1 & 0.99 & 0\% & 0\% & 1\% & 15\% & 43\%
\\\hline
5  & 1 & 1    & 0\% & 0\% & 1\% & 16\% & 43\%
\\\hline
6  & 1 & 0.99 & 0\% & 0\% & 1\% & 15\% & 41\%
\\\hline
\end{tabularx}
\end{table*}
\newpage
%\scenarioFigureCompact{153502}

\subsubsection{Guarantees}
Throughout the experiment, it was possible to guarantee the quality-of-solution for $\mathcal{P}_\textit{diagonal}$, by bounding $\textit{loss}(\mathcal{P},\mathcal{P}_\textit{diagonal})$ in post-solution evaluation -- after solving each (simplified) planning session, and before applying the selected action. Obviously no bound should be calculated for $\mathcal{P}_\textit{involved}$, since the loss was guaranteed to be zero in our pre-solution "offline" evaluation.
As explained in Section~\ref{sec:guarantees-general}, (\ref{eq:loss-bound-post-post}) provides a formula for the loss bound, given the solution of the simplified problem (which is available), and some domain-specific bounds/limits for the objective function. Here, we used the topological bounds from (\ref{eq:lb-topological}) and (\ref{eq:ub-topological}), and assigned them in the formula to provide guarantees during each planning session.

The tightness of these topological bounds, which affects the tightness of the loss bound, depends on the ratio between the angular variance, and the position variance, with which we model the noise in factors between poses; the smaller the angular noise is, in relation to the latter, the tighter the bounds are (as analyzed by \cite{Khosoussi18ijrr} and by \cite{Kitanov19arxiv}). Hence, we calculated the loss bound assuming different noise models (different such ratios), and examined their effects. Such a change to the noise model has a minor effect on the objective evaluation, since it does not change the sparsity pattern of the matrices; thus, we only present the effect on the inferred loss bound, and not on the entire planning process. The bounds, which were calculated assuming different noise ratios, are given in Table~\ref{tbl:scenario-loss}. The loss and its bounds are brought as a percentage of the maximal approximated objective function value in that session, to allow a correct comparison. In the scenario showcased before, the angular variance to position variance ratio was 0.25:1.

Indeed, changing the noise model has a significant influence on the tightness of the loss bounds. A ratio of 0.01:1 yields a very tight bound. It is not far-fetched that the angular variance would be this low in a navigation scenario, for example, by having a compass, as mentioned before.
Raising this ratio results in more conservative bounds, especially in comparison to the exact loss, which is zero. Yet they can still be used to guarantee that the solution stays in an acceptable range. Developing tighter bounding methods for the objective function shall help making these guarantees less conservative. 

To clarify, this discussion, alongside any assumptions on the noise or state structure, is only brought in order to examine our ability to provide guarantees, using this specific topological method. It is not essential in any way in order to apply the sparsification and improve the performance.

\section{Conclusions}
In an attempt to allow efficient autonomous decision making, and, specifically, decision making in the (high-dimensional) belief space, we introduced a new solution paradigm, which suggests performing a conscious simplification of the decision problem. Its impact is intended to be both conceptual and practical. 
Conceptually, we claimed that decision making, i.e., identification of the best candidate action, can utilize a simplified representation or approximation of the initial state, without compromising the accuracy of the state inference process. After efficiently selecting a candidate action, it should be applied on the \emph{original} state, which remains exact. On top of that, we presented the \emph{simplification loss} as a quality of solution measure, and explained how it can be bounded (e.g., using the \emph{simplification offset}) in order to provide guarantees. We recognized that when the simplification maintains \emph{action consistency}, i.e., when the trend of the objective function is maintained after the simplification, there is no loss. 

Practically, when applying the paradigm to the belief space, decision making can be conducted considering a sparse approximation of the prior belief. We provided a scalable algorithm for generation of such approximations. This versatile algorithm can generate approximations of different degrees, based on the subset of state variables selected for the sparsification. Specifically, by identifying the problem's \emph{uninvolved variables}, we can provide an action consistent approximation, which is \emph{guaranteed} to preserve the action selection. As explained in Section~\ref{sec:prob-analysis}, our sparsification approach is original and intuitive, as it exploits the belief's underlying Bayes net structure. We presented an in-depth study of our approach, and demonstrated it in a highly realistic active SLAM simulation. We showed that using sparsification of uninvolved variables, planning time can be significantly reduced, while, as mentioned, guaranteeing no loss in the quality of solution. We then showed that planning time can be reduced even further, when sparsifying all the state variables; in practice, for this configuration, we experienced no loss in the quality of solution, as well. Nonetheless, we demonstrated how the theoretical loss in that case can be bounded.

The proposed novel paradigm offers many possible future research directions. In general, other sparsification methods, besides the provided algorithm, can be used in similar ways; however, their impact on the action selection should be examined. Potentially, existing (approximated) solution methods for POMDPs can also be evaluated with our theoretical framework, to provide a standard comparison tool for measuring the accuracy of planning algorithms. Also, this framework can be used to develop a scheme for elimination of candidate actions; in fact, we have already developed a proof of concept for this idea \citep{Elimelech17isrr}. We can also examine other simplification methods, such as altering the action set or the objective function. Developing simplification methods for more general beliefs, such as multi-modal Gaussians, can hold important practical significance. Derivation of tighter loss bounds is also of interest. Overall, with the versatility of these ideas, we expect the approach to yield a substantial contribution to the research community.

\section{Acknowledgments}
The authors would like to acknowledge Dr.~Andrej Kitanov from the Faculty of Aerospace Engineering at the Technion –- Israel Institute of Technology, for insightful discussions concerning Section~\ref{sec:dmuu-guarantees}, and his assistance with implementing the simulation.

\section{Declaration of conflicting interest}
The authors declare that there is no conflict of interest.

\section{Funding}
This work was supported by the Israel Science Foundation (grant 351/15). 

\bibliography{../../../References/refs}

\begin{appendices}
\section{Additional loss bounds \label{apndx:bounds}}
We present here additional techniques to bound the loss between a decision problem $\mathcal{P} \doteq \left(b,\mathcal{U},J\right)$, and its simplified version $\mathcal{P}_s \doteq \left(b_s,\mathcal{U},J\right)$, which uses a sparse belief approximation, created with Algorithm~\ref{alg:sparsification}.

\subsection{Pre-solution guarantees: rank-1~updates \label{apndx:pre-eval-guarantees}}
We remind again that according to Lemmas~\ref{trm:loss-bound-pre}~and~\ref{trm:loss-bound-post} in Section~\ref{sec:guarantees-general}, we can use (a bound of) the offset between the problem and its simplification, to derive a loss bound. In Section~\ref{sec:dmuu-var-selection}, we proved that sparsification of the uninvolved variables always results in zero offset, and hence zero loss. Now, we show that under additional restrictions, we can derive an offset bound also when sparsifying involved variables.

Assume that for every action $u \in\mathcal{U}$ the corresponding collective Jacocian $\bm U \in\mathbb{R}^{1\times N}$ contains only a single row, i.e., rank-1 information updates. This can be the case, for example, in sensor placement problems with scalar measurements (like temperature). Now, let us analyze the simplification offset:
%\vspace{-20pt}
\begin{flushright}
\begin{align}
&2\cdot\delta(\mathcal{P},\mathcal{P}_s,u) = \\
&2\cdot\abs{V(b,u)-V(b_s,u)} = \\
&\abs{\ln\abs{\bm \Lambda+ \bm U^T \bm U} - \ln\abs{\bm \Lambda_s+ \bm U^T \bm U}} = \\
	\shortintertext{ (Matrix determinant lemma \citep[see][]{Harville98book_matrix})\hspace{1cm} }
&\lvert \ln\left(\abs{\bm \Lambda}\cdot\left(1+\bm U \bm \Lambda^{-1} \bm U^T\right)\right) -  \nonumber\\
& \hspace{3cm} \ln\left(\abs{\bm \Lambda_s}\cdot\left(1+\bm U \bm \Lambda^{-1}_s \bm U^T\right)\right) \rvert = \\
	\shortintertext{ (Eq.~\ref{eq:sparse-det-eq})\hspace{1cm} }
&\abs{ \ln\left(1+\bm U \bm \Lambda^{-1} \bm U^T\right) - \ln\left(1+\bm U \bm \Lambda^{-1}_s \bm U^T\right) } = \\
&\lvert \ln\left(1+\bm U \bm \Lambda^{-1}_s \bm U^T + \bm U (\bm \Lambda^{-1}-\bm \Lambda^{-1}_s) \bm U^T\right) -  \nonumber\\
& \hspace{4cm} \ln\left(1+\bm U \bm \Lambda^{-1}_s \bm U^T\right) \rvert = (\star)
\end{align}
\end{flushright}
The logarithm is a monotonously increasing concave function, thus, every $a,b \in \mathbb{R}$ and $c\geq0$ satisfy
\begin{equation}
\abs{\ln(a)-\ln(b)} \geq \abs{\ln(a+c) - \ln(b+c)}.
\end{equation}
In other words, the difference in the function value between a pair of inputs decreases, when the inputs equally grow.\linebreak
Surely, $0 \leq \bm U \bm \Lambda^{-1}_s \bm U^T$, since $\bm \Lambda^{-1}_s$ is positive semi-definite. Thus, we may choose $a = 1+ \bm U (\bm \Lambda^{-1}-\bm \Lambda^{-1}_s) \bm U^T,\, b = 1$, and $c = \bm U \bm \Lambda^{-1}_s \bm U^T $. Therefore,
\begin{align}
(\star) \leq &\abs{ \ln\left(1+ \bm U (\bm \Lambda^{-1}-\bm \Lambda^{-1}_s) \bm U^T\right) - \ln\left(1\right) } = \\
&\abs{ \ln\left(1+ \bm U (\bm \Lambda^{-1}-\bm \Lambda^{-1}_s) \bm U^T\right) } \leq \\
&\abs{ \ln\left(1+  \alpha \cdot\sum_{i,j\in\inv(u)}(\bm \Lambda^{-1}-\bm \Lambda^{-1}_s)_{ij}  \right) },
\end{align}
where $\inv(u)$ is the set of (prior state) variables involved in $u$, and the scalar $\alpha$ complies to $\alpha \geq \max_i \bm U_i^2 $. We recall that $\bm U_i$ is uninvolved $\iff \bm U_i=0$. When considering the involved variables among all the actions, and $\alpha$ is valid $\forall u\in \mathcal{U}$, this bound becomes independent of a specific action, and only a single expression needs to be calculated. Overall, we can conclude the following bound on the offset:
\begin{empheq}[box=\fbox]{multline}
\Delta(\mathcal{P},\mathcal{P}_s) \leq\\ \frac12\cdot\abs{ \ln\left(1+  \alpha \cdot\sum_{i,j\in\inv(\mathcal{U})} (\bm \Lambda^{-1}-\bm \Lambda^{-1}_s)_{ij}  \right) }. \label{eq:bound-ro-single-row}
\end{empheq}

As we may notice, this symbolic bound depends on the initial belief of the original and simplified problems, yet not on their solution; it hence can be utilized before actually solving the problem.
When calculating this bound, we considered only single-row collective Jacobians, but otherwise arbitrary.
Although, the considered assumption is restrictive, the concluded bound is indeed usable for certain problems, as evident in our follow-up work \citep{Elimelech17isrr}. Guaranteed action consistency for the case of single-row Jacobians, which are also limited to a single non-zero entry, was previously shown by \cite{Indelman16ral}.% Also, this derivation was principal in demonstrating the concept of "semi-online" bounds. %Certainly, deriving better and more general guarantees is an ongoing work.

\subsection{Post-solution guarantees \label{apndx:post-eval-guarantees}}
We recall that the offset can also be bounded by utilizing domain-specific upper and lower bounds of the objective function ($\mathcal{UB},\mathcal{LB}$, respectively), as indicated in (\ref{eq:offset-bound-post}).
In addition to the topological objective bounds, which were presented in Section~\ref{sec:dmuu-guarantees}, we may also utilize alternative bounds, which rely on known determinant bounds. 

For the lower bound, we can use \emph{Minkowski determinant inequality}, which states that for positive semi-definite matrices $\bm M_1, \bm M_2 \in \mathbb{R}^{N\times N}$  
\begin{align}
\abs{\bm M_1+\bm M_2}^\frac1N &\geq \abs{\bm M_1}^\frac1N + \abs{\bm M_2}^\frac1N, \\
\ln \abs{\bm M_1+\bm M_2} &\geq N \cdot \ln \left( \abs{\bm M_1}^\frac1N + \abs{\bm M_2}^\frac1N\right).
\end{align}
Let us assign $\bm M_1\doteq \bm \Lambda, \bm M_2\doteq \bm U^T \bm U$; when $\bm U^T \bm U$ is not a full rank update (e.g. $\bm U$ has less than $N$ rows), $\abs{\bm U^T \bm U}=0$, and we are left with
\begin{equation}
\ln\abs{\bm \Lambda+\bm U^T \bm U} \geq \ln \abs{\bm \Lambda} \label{eq:minkowski}
\end{equation}
For formality, it is easy to show that even if the prior state size is smaller than $N$, the validity of the conclusion is not compromised. % (in a similar way to Eq.~\ref{eq:lim-aug-det}).
For the upper bound, we can use \emph{Hadamard inequality}, which states that for a positive semi-definite matrix $\bm M \in \mathbb{R}^{N\times N}$ 
\begin{align}
\abs{\bm M} & \leq \prod_{i=1}^N (\bm M)_{ii}. \\
\intertext{Let us assign $\bm M\doteq\bm \Lambda + \bm U^T \bm U$; then}
\abs{\bm \Lambda+\bm U^T \bm U} &\leq \prod_{i=1}^N (\bm \Lambda+\bm U^T \bm U)_{ii}, \\
\ln\abs{\bm \Lambda+\bm U^T \bm U} &\leq \sum_{i=1}^n \ln [(\bm \Lambda+\bm U^T \bm U)_{ii}].
\end{align}
Overall, we get the following objective function bounds:
\begin{empheq}[box=\fbox]{align}
\mathcal{LB}_\text{det}\left\{ V(b,u) \right\} &\doteq \ln \abs{\bm \Lambda} - N\cdot\ln (2\pi e), \label{eq:lb-det} \\
\mathcal{UB}_\text{det}\left\{ V(b,u) \right\} &\doteq \sum_{i=1}^N \ln [(\bm \Lambda+\bm U^T \bm U)_{ii}]  \nonumber\\
& \hspace{2.5cm}  - N\cdot\ln (2\pi e), \label{eq:ub-det}
\end{empheq}
where $\bm \Lambda$ is the information matrix of prior belief $b$, and $\bm U$ is the collective Jacobian of action $u$, and $N$ is the posterior state size.

Unlike the bounds presented in Section~\ref{sec:dmuu-guarantees}, these bounds are extremely general, as they make no assumptions on the state nor actions, besides the standard problem formulation. As expected, this advantage comes at the expense of tightness. Nonetheless, they may especially be useful when the matrix $\bm \Lambda$ is diagonally dominant.

\pagebreak

\section{Proofs \label{apndx:proofs}}
\subsection{Lemma~\ref{trm:loss-bound-pre}}
\begin{proof} $ $\\
Refer to the proof of the more general case, stated in Lemma~\ref{trm:loss-bound-unbiased}.
\\\qed
\end{proof}

\subsection{Lemma~\ref{trm:loss-bound-post}}
\begin{proof} $ $\\
Refer to \cite{Elimelech21thesis} for an or an extended discussion and formulation of this statement.
\\\qed
\end{proof}

\subsection{Lemma~\ref{trm:ac-equiv-relation}}
The properties are trivially given from the definition of action consistency.
\\\qed

\subsection{Lemma~\ref{trm:ac-iff-f}}
\begin{proof} $ $\\
Assume $f$ is a monotonously increasing function such that for every two actions $a_i,a_j \in \mathcal{A}$
\begin{equation}
f(V_1({\bm\xi}_1,a_i)) = V_2({\bm\xi}_2,a_i),\quad f(V_1({\bm\xi}_1,a_j)) = V_2({\bm\xi}_2,a_j),
\end{equation}
then
\begin{multline}
f(V_1({\bm\xi}_1,a_i)) < f(V_1({\bm\xi}_1,a_j))\iff \\ V_2({\bm\xi}_2,a_i) < V_2({\bm\xi}_2,a_j),
\end{multline}
Because $f$ is monotonously increasing, then $f(x) < f(y)$ $ \iff x < y$, and 
\begin{equation}
\begin{gathered}
V_1({\bm\xi}_1,a_i) < V_1({\bm\xi}_1,a_j)\iff V_2({\bm\xi}_2,a_i) < V_2({\bm\xi}_2,a_j)%, \\
%V_1({\bm\xi}_1,a_i) = V_1({\bm\xi}_1,a_j)\iff V_2({\bm\xi}_2,a_i) = V_2({\bm\xi}_2,a_j)
. \\
\end{gathered}
\end{equation}
Meaning, $\left({\bm\xi}_1,\mathcal{A},V_1\right) \simeq \left({\bm\xi}_2,\mathcal{A},V_2\right)$.

Now to prove the opposite direction, assume $\left({\bm\xi}_1,\mathcal{A},J_1\right) \simeq \left({\bm\xi}_2,\mathcal{A},J_2\right)$; hence,
\begin{equation}
\label{eq:ac-conditions}
\begin{aligned}
V_1({\bm\xi}_1,a_i) < V_1({\bm\xi}_1,a_j)\iff V_2({\bm\xi}_2,a_i) < V_2({\bm\xi}_2,a_j)%, \\
%V_1({\bm\xi}_1,a_i) = V_1({\bm\xi}_1,a_j)\iff V_2({\bm\xi}_2,a_i) = V_2({\bm\xi}_2,a_j)
.
\end{aligned}
\end{equation}
Let us define a new function $f$ on the domain $\left\{ V_1({\bm\xi}_1,a) \mid a\in\mathcal{A} \right\}$ such that $f(V_1({\bm\xi}_1,a)) \doteq V_2({\bm\xi}_2,a)$. Given this definition and the action consistency conditions from (\ref{eq:ac-conditions}), we can conclude that
\begin{multline}
f(V_1({\bm\xi}_1,a_i)) < f(V_1({\bm\xi}_1,a_j)) \iff \\ V_2({\bm\xi}_2,a_i) < V_2({\bm\xi}_2,a_j) \iff \\ V_1({\bm\xi}_1,a_i) < V_1({\bm\xi}_1,a_j).
\end{multline}
Thus, $f$ is monotonously increasing on its domain.
\\\qed
\end{proof}

\subsection{Lemma~\ref{trm:ro-iff-ac}}
\begin{proof} $ $\\
Both directions are a direct consequence of Lemma~\ref{trm:ac-iff-f}.
Assume $\Delta^*(\mathcal{{\bm P}},\mathcal{{\bm P}}_s) = 0$. Thus, a monotonously increasing function $f$ exists such that $\Delta(\mathcal{{\bm P}},\mathcal{{\bm P}}^f_s) = 0$. Meaning, for every action $a \in \mathcal{A}$, $f(V_s({\bm\xi}_s,a))=V({\bm\xi},a)$. According to Lemma~\ref{trm:ac-iff-f}, it is sufficient to prove that $\mathcal{{\bm P}} \simeq \mathcal{{\bm P}}_s$.

To prove the opposite direction, assume $\mathcal{{\bm P}} \simeq \nolinebreak\mathcal{{\bm P}}_s$. Let us define a new function $f$ on the domain $\left\{ V_s({\bm\xi}_s,a) \mid a\in\mathcal{A} \right\}$ such that $f(V_s({\bm\xi}_s,a)) \doteq V({\bm\xi},a)$. From this definition, $\Delta(\mathcal{{\bm P}},\mathcal{{\bm P}}^f_s) = 0$. Also, according to Lemma~\ref{trm:ac-iff-f}, this function~$f$ is monotonously increasing, and thus $\Delta^*(\mathcal{{\bm P}},\mathcal{{\bm P}}_s) = 0$.
\\\qed
\end{proof}

\subsection{Lemma~\ref{trm:loss-bound-unbiased}}
\begin{proof} $ $\\
From the definition of the simplification offset, we know that for every monotonously increasing function $f$, the following is true:
\begin{align}
\abs{V({\bm\xi},a^*) - f(V_s({\bm\xi}_s,a^*))} &\leq \Delta(\mathcal{P},\mathcal{P}^f_s), \\
\abs{V({\bm\xi},a^*_s) - f(V_s({\bm\xi}_s,a^*_s))} &\leq \Delta(\mathcal{P},\mathcal{P}^f_s).
\end{align}
Removing the absolute values surely does not compromise the inequalities:
\begin{align}
V({\bm\xi},a^*) - f(V_s({\bm\xi}_s,a^*)) &\leq \Delta(\mathcal{P},\mathcal{P}^f_s), \\
f(V_s({\bm\xi}_s,a^*_s)) - V({\bm\xi},a^*_s) &\leq \Delta(\mathcal{P},\mathcal{P}^f_s).
\end{align}
By adding the two inequalities, and utilizing the definition of the $\loss$, we get:
\begin{multline}
\loss(\mathcal{P},\mathcal{P}^f_s) + f(V_s({\bm\xi}_s,a^*_s)) - f(V_s({\bm\xi}_s,a^*)) \\\leq 2\cdot \Delta(\mathcal{P},\mathcal{P}^f_s).
\end{multline}
From the definition of $a^*_s$, we know that
\begin{equation}
V_s({\bm\xi}_s,a^*_s)) \geq V_s({\bm\xi}_s,a^*).
\end{equation}
Since $f$ is monotonously increasing, then also
\begin{align}
f(V_s({\bm\xi}_s,a^*_s))) &\geq f(V_s({\bm\xi}_s,a^*)), \\
f(V_s({\bm\xi}_s,a^*_s))) - f(V_s({\bm\xi}_s,a^*)) &\geq 0.
\end{align}
Thus, we can infer that
\begin{equation}
\loss(\mathcal{P},\mathcal{P}^f_s) \leq 2\cdot \Delta(\mathcal{P},\mathcal{P}^f_s).
\end{equation}

Since the final statement is true for any monotonously increasing function $f$, we may conclude the desired upper bound over the loss, 
\begin{equation}
\loss(\mathcal{{\bm P}},\mathcal{{\bm P}}_s) \leq 2\cdot\Delta^*(\mathcal{{\bm P}},\mathcal{{\bm P}}_s)
\end{equation}
\qed
\end{proof}

\subsection{Lemma~\ref{trm:ro-triang}}
\begin{proof} $ $\\
Let us examine three decision problems $\mathcal{{\bm P}}_1,\mathcal{{\bm P}}_2,\mathcal{{\bm P}}_3$, where $\mathcal{P}_i \doteq  \left({\bm\xi}_i,\mathcal{A},V_i\right)$.
First, let us define the notation $ \delta(\mathcal{{\bm P}}_i,\mathcal{{\bm P}}_j,a) \doteq \abs{V_i({\bm\xi}_i,a) - V_j({\bm\xi}_j,a)} $. Now, for each two problems $\mathcal{{\bm P}}_i,\mathcal{{\bm P}}_j$, we mark $a_{ij}\in\mathcal{A}$ as the action, and $f_{ij}$ as the balance function, for which $\Delta^*(\mathcal{{\bm P}}_i,\mathcal{{\bm P}}_j) \doteq \delta(\mathcal{{\bm P}}_i,\mathcal{{\bm P}}^{f_{ij}}_j,a_{ij})$ (the values can be chosen arbitrarily from all values which comply to the equation). According to this notation we can conclude:
\begin{equation}
\begin{gathered}
\Delta^*(\mathcal{{\bm P}}_1,\mathcal{{\bm P}}_2) + \Delta^*(\mathcal{{\bm P}}_2,\mathcal{{\bm P}}_3) \doteq \\
\delta(\mathcal{{\bm P}}_1,\mathcal{{\bm P}}^{f_{12}}_2,a_{12}) + \delta(\mathcal{{\bm P}}_2,\mathcal{{\bm P}}^{f_{23}}_3,a_{23}) \geq \\
\delta(\mathcal{{\bm P}}_1,\mathcal{{\bm P}}^{f_{12}}_2,a_{13}) + \delta(\mathcal{{\bm P}}_2,\mathcal{{\bm P}}^{f_{23}}_3,a_{13}) \doteq \\
\abs{V_1({\bm\xi}_1,a_{13}) - f_{12}(V_2({\bm\xi}_2,a_{13}))} + \hspace{3cm}\\ \hfill\abs{V_2({\bm\xi}_2,a_{13}) - f_{23}(V_3({\bm\xi}_3,a_{13}))} \geq \\
\vert V_1({\bm\xi}_1,a_{13}) - f_{12}(V_2({\bm\xi}_2,a_{13})) + \hfill\\ \hfill V_2({\bm\xi}_2,a_{13}) - f_{23}(V_3({\bm\xi}_3,a_{13}))\vert \doteq (\star\star).
\end{gathered}
\end{equation}

Let us define the following scalar function:
\begin{multline}
F(x) \doteq f_{23}(x) + f_{12}(V_2({\bm\xi}_2,a_{13})) - V_2({\bm\xi}_2,a_{13}) = \\ f_{23}(x) + \text{constant}.
\end{multline}
Since $f_{23}$ is a monotonously increasing, so is $F$, and
\begin{equation}
\begin{gathered}
(\star\star) = \abs{V_1({\bm\xi}_1,a_{13}) - F(V_3({\bm\xi}_3,a_{13}))} \doteq \\
\delta(\mathcal{{\bm P}}_1,\mathcal{{\bm P}}^F_3,a_{13}) \geq \\
\delta(\mathcal{{\bm P}}_1,\mathcal{{\bm P}}^{f_{13}}_3,a_{13}) = \\
\Delta^*(\mathcal{{\bm P}}_1,\mathcal{{\bm P}}_3).
\end{gathered}
\end{equation}
Hence, $\Delta^*$ satisfies the triangle inequality.
\\\qed
\end{proof}

\subsection{Corollary~\ref{cor:reordering-in-root}}
\begin{proof} $ $\\
Let us mark as $\bm R^p_s$ the sparsified square root matrix, before permuting the variables back to their original order in line~\ref{alg:sparsification-line:permute-back} of Algorithm~\ref{alg:sparsification}. First, we show that applying the reverse permutation $\bm P \Box \bm P^T$ on $\bm R^p_s$ indeed leads to a square root of the sparse information matrix $\bm \Lambda_s$ (in the original order):
\begin{equation}
 (\bm P \bm R^p_s \bm P^T)^T(\bm P \bm R^p_s \bm P^T) = \bm P {\bm R^p_s}^T \bm R^p_s \bm P^T = \bm P \bm \Lambda^p_s \bm P^T  = \bm \Lambda_s,
\end{equation}
where $\bm \Lambda^p_s$ is the sparsified information matrix, before permuting the variables back.

Now, we want to examine the shape of the matrix $ \bm R_s \doteq \bm P \bm R^p_s \bm P^T $, and show that it is indeed triangular. According to Algorithm~\ref{alg:sparsification}, before executing line~\ref{alg:sparsification-line:permute-back}, $\bm R^p_s$ is of the following structure:
\begin{equation}
\bm R^p_s =
\left(\begin{array}{c|c}
\text{diagonal} & \bm 0 \\ \hline \bm 0 & \text{triangular}
\end{array}\right),
\end{equation}
where the rows of the diagonal block correspond to the sparsified variables. Without losing generality, we should only prove that applying a permutation of the form $p'\mathpunct{:}\,(1,\dots,n)\mapsto(2,\dots,i,1,i+1,\dots,n)$ on this matrix (i.e., "pushing forwards" one of the sparsified variables), does not break the triangular form.
Hence, assuming $\bm P^T$ is the column permutation matrix matching such $p'$, let us look at

\begin{align}
\bm R_s \doteq \bm P \bm R^p_s \bm P^T = 
\nonumber\\
& \bm P \left(\begin{array}{c|c}
d \in \mathbb{R} & 0\dots0 \\ \hline \begin{array}{c} 0 \\ \vdots \\ 0 \end{array}  & \text{triangular}
\end{array}\right) \bm P^T =
\nonumber\\
&\left(\begin{array}{c|c|c}
\begin{array}{c} 0 \\ \vdots \\ 0 \end{array}  & \text{triangular} & \bm\ast \\ \hline d & 0\dots0 & 0\dots0\\ \hline \begin{array}{c} 0 \\ \vdots \\ 0 \end{array} & \bm 0 & \text{triangular}
\end{array}\right) \bm P^T =
\nonumber \\
&\left(\begin{array}{c|c|c}
\text{triangular} & \begin{array}{c} 0 \\ \vdots \\ 0 \end{array}  & \bm\ast \\ \hline 0\dots0 & d & 0\dots0\\ \hline \bm 0 & \begin{array}{c} 0 \\ \vdots \\ 0 \end{array}  & \text{triangular}
\end{array}\right).
\end{align}
Recursively utilizing this conclusion, for more intricate permutations, proves that $\bm R_s$ is indeed triangular, whenever permuting the sparsified variables back to their original order, as desired.
\\\qed
\end{proof}

\subsection{Theorem~\ref{trm:uninvolved-is-ac}}
\begin{proof} $ $\\
Consider a belief $b = \mathcal{N}({\bm X}^*,\bm \Lambda^{-1})$, where the state contains $n_1$ uninvolved variables and $n_2$ involved variables, such that $n = n_1 + n_2$ is the prior state size. Also consider the simplified belief $b_s = \mathcal{N}({\bm X}^*,\bm \Lambda_s^{-1})$, in which all uninvolved variables were sparsified, by applying Algorithm~\ref{alg:sparsification}.

We mark with $\bm P$ the (column) permutation matrix that positions all the involved variable at the end of the state. Now, let ${\bm R}^p$ be the Cholesky factor of the permuted information matrix ${\bm \Lambda}^p \doteq {\bm P}^T {\bm \Lambda} {\bm P}$, such that ${\bm \Lambda}^p ={{\bm R}^p}^T {\bm R}^p$. This ${\bm R}^p$ can be divided into block form:
\begin{equation}
{\bm R}^p \doteq 
\left(\begin{array}{c|c}
{\bm R}^p_{11} & {\bm R}^p_{12} \\ \hline \bm 0^{n_2 \times n_1} & {\bm R}^p_{22}
\end{array}\right),
\end{equation}
where ${\bm R}^p_{11} \in\mathbb{R}^{n_1 \times n_1}$ and ${\bm R}^p_{22} \in\mathbb{R}^{n_2 \times n_2}$ are triangular sub-matrices, ${\bm R}^p_{12} \in\mathbb{R}^{n_1 \times n_2}$, and $\bm 0^{n_1 \times n_2}$ is a zero matrix in the specified size.
By following the steps of Algorithm~\ref{alg:sparsification}, we realize that the returned sparsified information matrix~$\bm \Lambda_s$ is given as $\bm \Lambda_s \doteq {\bm P} {{\bm R}^p_s}^T {\bm R}^p_s{\bm P}^T$ (or, equally, satisfies \mbox{ ${\bm P}^T \bm \Lambda_s {\bm P} \doteq {{\bm R}^p_s}^T {\bm R}^p_s$}), where 
\begin{equation}
{\bm R}^p_s \doteq 
\left(\begin{array}{c|c}
{\bm D}^p_{11} & \bm 0^{n_1 \times n_2} \\ \hline \bm 0^{n_2 \times n_1} & {\bm R}^p_{22}
\end{array}\right),
\end{equation}
and ${\bm D}^p_{11}$ is the diagonal matrix formed by copying the diagonal of ${\bm R}^p_{11}$ (and assigning zero elsewhere).

We would like to find the simplification offset between the two decision problems $\problem$ and $\problem_s$ (for which $b$ and $b_s$ are the initial beliefs, respectively). Let us consider a candidate action $u\in\mathcal{{\bm U}}$ with a collective Jacobian ${\bm U}\in\mathbb{R}^{h\times (n+m)}$, where $n+m$ is the posterior state size.
We may derive the following from the definition of the offset and the objective function $V$:
\begin{equation}
\label{eq:proof-trm-offset-of-pi}
\delta(\problem,\problem_s,u) = \frac12\cdot\abs{ \ln \abs{\breve{{\bm \Lambda}} + \bm U^T \bm U} - \ln \abs{\breve{{\bm \Lambda}_s} + \bm U^T \bm U}}.
\end{equation}

Now, let us examine the following expression:
\begin{equation}
\# \doteq \abs{\breve{{\bm \Lambda}} + \bm U^T \bm U} - \abs{\breve{{\bm \Lambda}}_s + \bm U^T \bm U},
\end{equation}

We know that (unitary) variable permutation does not affect the determinant of a matrix, thus
\begin{align}
\# = \abs{\breve{\bm P}^T\left(\breve{{\bm \Lambda}} + \bm U^T \bm U \right)\breve{\bm P}} &- \abs{\breve{\bm P}^T\left(\breve{{\bm \Lambda}}_s + \bm U^T \bm U \right)\breve{\bm P}}
= \nonumber \\
\abs{\breve{\bm P}^T\breve{\bm \Lambda}\breve{\bm P} + (\bm U \breve{\bm P})^T (\bm U \breve{\bm P}) } &- 
\abs{\breve{\bm P}^T\breve{{\bm \Lambda}}_s\breve{\bm P} + (\bm U \breve{\bm P})^T (\bm U \breve{\bm P})},
\end{align}
where
\begin{equation}
\breve{\bm P}\doteq 
\left(\begin{array}{c|c}
\bm P & \bm 0^{n \times m} \\ \hline \bm 0^{m \times n} & \bm I^{m \times m}
\end{array}\right)
\end{equation}
is the augmented permutation matrix, which keeps the variables added in the update at the end of the state. Note that if the variables were not originally added to the end of the state, the permutation $\breve{\bm P}$ can be easily adapted to enforce this property.

We can also augment the matrix ${\bm R}^p$ with $m$ empty columns (and similarly for ${\bm R}^p_s$):
\begin{equation}
\breve{{\bm R}^p} \doteq 
\left(\begin{array}{c|c|c}
{\bm R}^p_{11} & {\bm R}^p_{12} &  \multirow{2}{*}{$\bm 0^{n \times m}$} 
\\ \cline{1-2} 
\bm 0^{n_2 \times n_1} & {\bm R}^p_{22} 
\end{array}\right),
\end{equation}
and assign the result in $\#$, to yield:
\begin{multline}
\# = \\
\abs{\breve{{\bm R}^p}^T \breve{{\bm R}^p} + (\bm U \breve{\bm P})^T (\bm U \breve{\bm P}) } - 
\abs{\breve{{\bm R}^p_s}^T \breve{{\bm R}^p_s} + (\bm U \breve{\bm P})^T (\bm U \breve{\bm P})}
\end{multline}
This expression can be reorganized to the following form:
\begin{equation}
\label{eq:trm-proof-delta-matrix}
\# = 
\abs{\begin{pmatrix} \breve{{\bm R}^p} \\ \bm U \breve{\bm P} \end{pmatrix}^T \begin{pmatrix} \breve{{\bm R}^p} \\ \bm U \breve{\bm P} \end{pmatrix}} - 
\abs{\begin{pmatrix} \breve{{\bm R}^p_s} \\ \bm U \breve{\bm P} \end{pmatrix}^T \begin{pmatrix} \breve{{\bm R}^p_s} \\ \bm U \breve{\bm P} \end{pmatrix}}
\end{equation}
The two matrices which appear in this expression also follow a block form:
\begin{align}
\begin{pmatrix} \breve{{\bm R}^p} \\ \bm U \breve{\bm P} \end{pmatrix} &= 
\left(\begin{array}{c|c|c}
{\bm R}^p_{11} & {\bm R}^p_{12} & \bm 0^{n_1 \times m}
\\ \cline{1-3} 
\bm 0^{(n_2+h) \times n_1} & \multicolumn{2}{c}{{\bm B}} 
\end{array}\right), \\
\begin{pmatrix} \breve{{\bm R}^p_s} \\ \bm U \breve{\bm P} \end{pmatrix} &= 
\left(\begin{array}{c|c}
{\bm D}^p_{11} & \bm 0^{n_1 \times (n_2+m)}
\\ \cline{1-2} 
\bm 0^{(n_2+h) \times n_1} & {\bm B} 
\end{array}\right),
\end{align}
where
\begin{equation}
{\bm B} \doteq 
\left(\begin{array}{c|c}
{\bm R}^p_{22} & \bm 0^{n_2 \times m}
\\ \hline
\multicolumn{2}{c}{{\bm U}^\text{inv}} 
\end{array}\right),
\end{equation}
and ${\bm U}^\text{inv}$ is a sub-matrix of $\bm U \breve{\bm P}$, containing its right \mbox{$n_2+m$} columns. Since the left $n_1$ columns of $\bm U \breve{\bm P}$ correspond to uninvolved variables, we know they may only contain zeros.

Thus, if we mark $\breve{{\bm R}^p_{12}} \doteq \left(\begin{array}{c|c} {\bm R}^p_{12} & \bm 0^{n_1 \times m} \end{array}\right)$, then the left term in (\ref{eq:trm-proof-delta-matrix}) is:
\begin{equation}
\abs{ \begin{pmatrix} \breve{{\bm R}^p} \\ \bm U \breve{\bm P} \end{pmatrix}^T \begin{pmatrix} \breve{{\bm R}^p} \\ \bm U \breve{\bm P} \end{pmatrix} }
=
\abs{ \begin{array}{c|c}
{{\bm R}^p_{11}}^T {\bm R}^p_{11} & {{\bm R}^p_{11}}^T \breve{{\bm R}^p_{12}} \\ \hline \breve{{\bm R}^p_{12}}^T {\bm R}^p_{11} & \breve{{\bm R}^p_{12}}^T \breve{{\bm R}^p_{12}} + {\bm B}^T {\bm B}
\end{array} }.
\end{equation}
From the block-determinant formula \citep[see][]{Harville98book_matrix}, this equals to
\begin{multline}
\abs{ {{\bm R}^p_{11}}^T {\bm R}^p_{11} } \cdot
\left\vert \breve{{\bm R}^p_{12}}^T \breve{{\bm R}^p_{12}} + {\bm B}^T {\bm B} - \dots\right.\\\left. \breve{{\bm R}^p_{12}}^T {{\bm R}^p_{11}} {{\bm R}^p_{11}}^{-1} {{{\bm R}^p_{11}}^T}^{-1}{{\bm R}^p_{11}}^T \breve{{\bm R}^p_{12}} \right\vert
 \\ = \abs{{{\bm R}^p_{11}}}^2 \cdot \abs{{\bm B}^T {\bm B}}
\end{multline}

The right term in (\ref{eq:trm-proof-delta-matrix}) is:
\begin{multline}
\abs{ \begin{pmatrix} \breve{{\bm R}^p_s} \\ \bm U \breve{\bm P} \end{pmatrix}^T \begin{pmatrix} \breve{{\bm R}^p_s} \\ \bm U \breve{\bm P} \end{pmatrix} }
=
\abs{ \begin{array}{c|c}
{{\bm D}^p_{11}}^T {\bm D}^p_{11} & {\bm 0} \\ \hline {\bm 0} & {\bm B}^T {\bm B}
\end{array} } \\
= 
\abs{ {{\bm D}^p_{11}}^2 } \cdot \abs{{\bm B}^T {\bm B} }
\end{multline}

Since ${\bm R}^p_{11}$ and ${\bm D}^p_{11}$ are triangular matrices with the same diagonal, their determinants are equal (to the product of the diagonal elements). Thus, $\# = 0$, and overall
\begin{equation}
\abs{ \breve{\bm \Lambda}+{\bm U}^T{\bm U} } = \abs{ \breve{\bm \Lambda}_s+{\bm U}^T{\bm U} }.
\end{equation}
This surely means that
\begin{equation}
\ln\abs{ \breve{\bm \Lambda}+{\bm U}^T{\bm U} } - \ln\abs{ \breve{\bm \Lambda}_s+{\bm U}^T{\bm U} } = 0.
\end{equation}
Finally, assigning this expression in (\ref{eq:proof-trm-offset-of-pi}) means that
\begin{equation}
\delta(\problem,\problem_s,u) = 0.
\end{equation}
Since the previous conclusion is true $\forall u\in\mathcal{U}$, this means that 
\begin{equation}
\Delta(\problem,\problem_s) \doteq \max_{u\in\mathcal{U}} \delta(\problem,\problem_s,u) = 0,
\end{equation}
as desired.
\\\qed
\end{proof}

\end{appendices}

\end{document}